\documentclass{article}

\PassOptionsToPackage{numbers, sort&compress}{natbib}

\usepackage[preprint]{neurips_data_2024}

\usepackage[utf8]{inputenc} %
\usepackage[T1]{fontenc}    %
\usepackage{hyperref}       %
\usepackage{url}            %
\usepackage{booktabs}       %
\usepackage{amsfonts}       %
\usepackage{nicefrac}       %
\usepackage{microtype}      %

\newcommand{\appref}[1]{\hyperref[#1]{Appendix~\ref*{#1}}}

\usepackage{graphicx}
\usepackage{booktabs}
\usepackage{multirow}
\usepackage{multicol}
\usepackage{indentfirst}
\usepackage{subcaption}
\usepackage{makecell}
\usepackage{mdframed}

\usepackage[ruled,vlined]{algorithm2e}
\usepackage[dvipsnames]{xcolor}
\definecolor{mygreen}{RGB}{58, 130, 51}
\definecolor{myblue}{RGB}{40, 84, 156}
\definecolor{mygray}{RGB}{142, 142, 142}
\definecolor{commentcolor}{RGB}{59,116,116}   %

\newcolumntype{M}[1]{>{\centering\raggedright\arraybackslash}m{#1}}
\usepackage{pifont}   %
\newcommand{\cmark}{\ding{51}} %
\newcommand{\xmark}{\textcolor{red}{\ding{55}}} %

\usepackage{nicematrix}
\definecolor{mygray}{gray}{0.85}
\definecolor{softgray}{rgb}{0.9, 0.9, 0.9}
\definecolor{softblue}{rgb}{0.88, 0.92, 1.0}
\definecolor{softgreen}{rgb}{0.88, 1.0, 0.88}
\definecolor{softyellow}{rgb}{1.0, 1.0, 0.88}
\definecolor{softred}{rgb}{1.0, 0.88, 0.88}
\definecolor{softpink}{rgb}{1.0, 0.88, 0.94}

\usepackage{color-edits}

\addauthor{gn}{magenta}
\addauthor{zq}{magenta}

\title{GenAI-Bench: Evaluating and Improving Compositional Text-to-Visual Generation}

\author{%
  Baiqi Li$^{1}$\thanks{Co-first authors; $^{\dag}$Co-senior authors. \href{https://linzhiqiu.github.io/papers/genai_bench}{Datasets and code are open-sourced on our website.} This manuscript extends our Best Short Paper at CVPR SynData24 Workshop~\cite{li2024genaibench}.} \quad Zhiqiu Lin$^{1,2*}$ \quad Deepak Pathak$^{1}$ \quad Jiayao Li$^{1}$ \quad Yixin Fei$^1$ \quad Kewen Wu$^1$ \\ 
  {\bf Tiffany Ling$^{1}$} \quad {\bf Xide Xia$^{2\dag}$} \quad {\bf Pengchuan Zhang$^{2\dag}$}  \quad {\bf Graham Neubig$^{1\dag}$} \quad {\bf Deva Ramanan$^{1\dag}$}  \\ 
  $^1$Carnegie Mellon University \quad\quad $^2$Meta 
}

\begin{document}

\maketitle

\begin{abstract}
  While text-to-visual models now produce photo-realistic images and videos, they struggle with {\em compositional} text prompts involving attributes, relationships, and higher-order reasoning such as logic and comparison. In this work, we conduct an extensive human study on {\bf GenAI-Bench} to evaluate the performance of leading image and video generation models in various aspects of compositional text-to-visual generation. We also compare automated evaluation metrics against our collected human ratings and find that {\bf VQAScore} -- a metric measuring the likelihood that a VQA model views an image as accurately depicting the prompt -- significantly outperforms previous metrics such as CLIPScore. In addition, VQAScore can improve generation in a black-box manner (without finetuning) via simply ranking a few (3 to 9) candidate images. {\bf Ranking by VQAScore} is 2x to 3x more effective than other scoring methods like PickScore, HPSv2, and ImageReward at improving human alignment ratings for DALL-E 3 and Stable Diffusion, especially on compositional prompts that require advanced visio-linguistic reasoning. We release a new {\bf GenAI-Rank} benchmark with over 40,000 human ratings to evaluate scoring metrics on ranking images generated from the same prompt. Lastly, we discuss promising areas for improvement in VQAScore, such as addressing fine-grained visual details. We will release all human ratings (over 80,000) to facilitate scientific benchmarking of both generative models and automated metrics.
\end{abstract}

\begin{figure*}[ht]
\centering
    \includegraphics[width=0.99\textwidth]{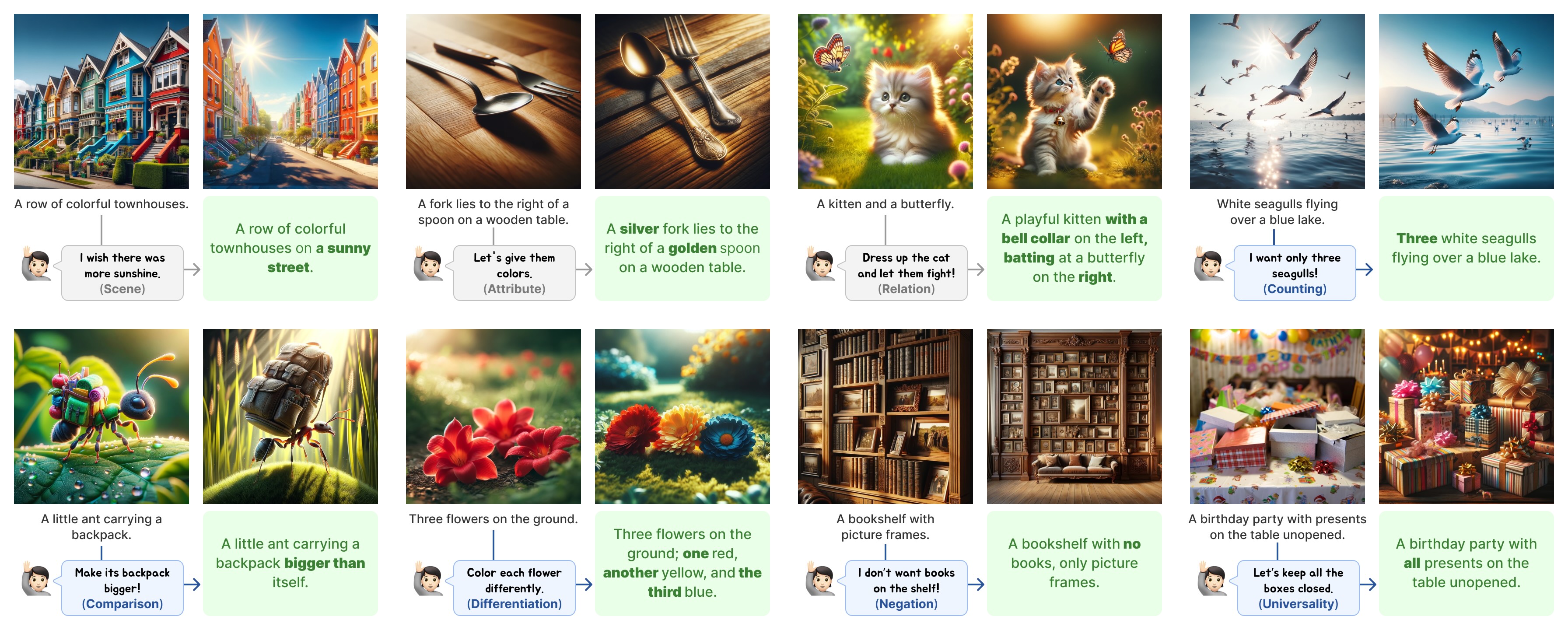}
    \caption{\small {\bf Compositional text prompts of our GenAI-Bench (highlighted in \textcolor{mygreen}{green})} reflect how real-world users (e.g., designers) seek precise control in text-to-visual generation. For example, users often write detailed prompts that specify compositions of basic visual entities and properties (highlighted in \textcolor{gray}{gray}), such as scenes, attributes, and relationships (spatial/action/part). Moreover, user prompts may require advanced visio-linguistic reasoning (highlighted in \textcolor{myblue}{blue}), such as counting, comparison, differentiation, and logic (negation/universality). \appref{sec:skill_details} provides skill definitions with more examples. \autoref{tab:skill_comparison} compares GenAI-Bench with previous benchmarks~\cite{parti, huang2023t2i, imagen, pickscore}. %
    }
    \label{fig:teaser}
\end{figure*}

\begin{figure*}[ht]
\centering
    \includegraphics[width=0.99\textwidth]{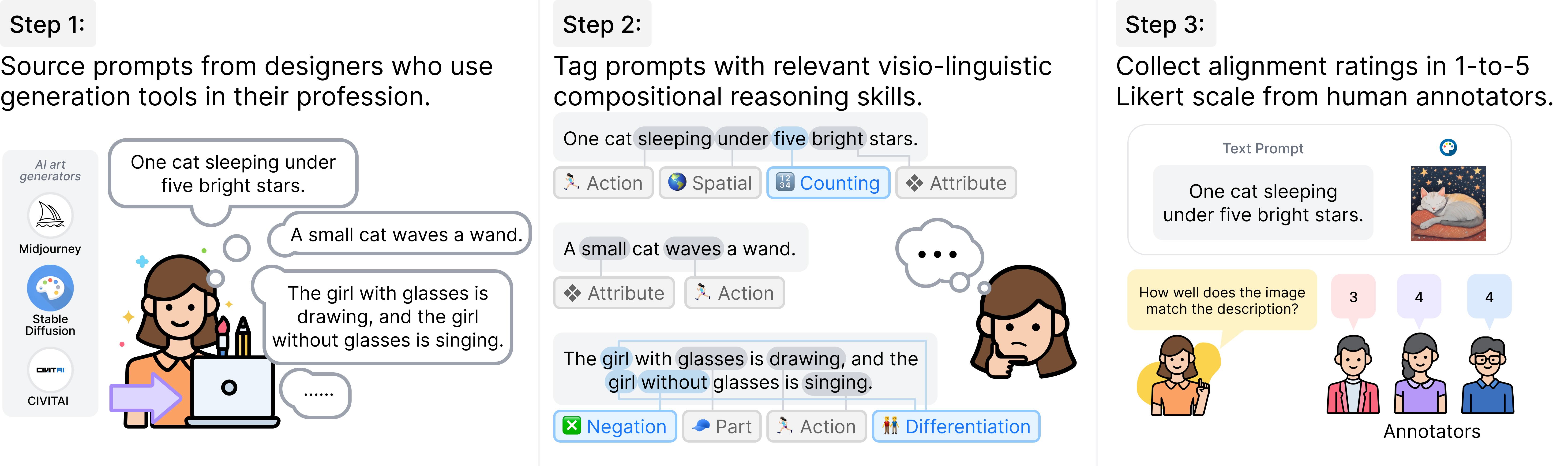}
    \caption{\small {\bf Collecting GenAI-Bench.} To reliably evaluate generative models, we source prompts from professional designers who use tools such as Midjourney~\cite{midjourney} and CIVITAI. This ensures the prompts encompass practical skills relevant to real-world applications and are free of subjective or inappropriate content. Each GenAI-Bench prompt is carefully tagged with all its evaluated skills. We then generate images and videos using state-of-the-art models like SD-XL~\cite{stablediffusion} and Gen2~\cite{gen2}. We follow the recommended annotation protocol~\cite{otani2023toward} to collect 1-to-5 Likert scale ratings for how well the generated visuals align with the input text prompts.
    }
    \label{fig:collection}
\end{figure*}

\section{Introduction}

State-of-the-art text-to-visual models like Stable Diffusion~\cite{stablediffusion}, DALL-E 3~\cite{dalle3}, Gen2~\cite{gen2}, and Sora~\cite{sora} generate images and videos with exceptional realism and quality. Due to their rapid advancement, traditional evaluation metrics and benchmarks (e.g., FID scores on COCO~\cite{FID, coco} and CLIPScores on PartiPrompt~\cite{clipscore, parti}) are becoming insufficient~\cite{otani2023toward, lin2024evaluating}. For instance, benchmarks should include more {\em compositional} text prompts~\cite{liu2022compositional} that involve attribute bindings, object relationships, and logical reasoning, among other visio-linguistic reasoning skills (\autoref{fig:teaser}). Moreover, it's crucial for automated evaluation metrics to measure how well the generated images (or videos) {\em align} with such compositional text prompts. Yet, widely used metrics like CLIPScore~\cite{clipscore} function as {\em bag-of-words}~\cite{winoground, aro, lin2024revisiting} and cannot produce reliable alignment (faithfulness~\cite{tifa}) scores. Therefore, to guide the scientific benchmarking of generative models, we conduct a comprehensive evaluation of compositional text-to-visual generation alongside automated alignment metrics~\cite{clipscore, davidsonian}.

{\bf Evaluating text-to-visual generation.} We evaluate generative models using our collected {\bf GenAI-Bench} benchmark~\cite{lin2024evaluating}, which consists of 1,600 challenging real-world text prompts sourced from professional designers. Compared to benchmarks~\cite{pickscore, tifa, evalcrafter} such as PartiPrompt~\cite{parti} and T2I-CompBench~\cite{huang2023t2i} (see \autoref{tab:skill_comparison}), GenAI-Bench captures a wider range of aspects in compositional text-to-visual generation~\cite{li2024evaluating}, ranging from {\bf basic} (scene, attribute, relation) to {\bf advanced} (counting, comparison, differentiation, logic). We collect a total of 38,400 human alignment ratings (1-to-5 Likert scales~\cite{otani2023toward}) on images and videos generated by ten leading models\footnote{In \cite{lin2024evaluating}, we collected 15,810 human ratings on a subset of 527 prompts. This paper extends it to 38,400 human ratings on all 1,600 prompts.}, such as Stable Diffusion~\cite{stablediffusion}, DALL-E 3~\cite{dalle3}, Midjourney v6~\cite{midjourney}, Pika v1~\cite{pikalab}, and Gen2~\cite{gen2}. \autoref{fig:collection} illustrates the evaluation process. Our human study shows that while these models can often accurately generate basic compositions (e.g., attributes and relations), they still struggle with advanced reasoning (e.g., logic and comparison). For instance, for ``basic'' prompts that do not require advanced reasoning, the state-of-the-art DALL-E 3 (most preferred by humans) achieves a remarkable average rating of $4.3$, meaning its images range from having ``{\em a few minor discrepancies}'' to `{\em matching exactly}'' with the prompts. However, its rating on ``advanced'' prompts drops to $3.4$, indicating ``{\em several discrepancies}''.  %

{\bf Evaluating automated metrics.} We also use the human ratings to benchmark automated metrics (e.g., CLIPScore~\cite{clipscore}, PickScore~\cite{pickscore}, and Davidsonian~\cite{davidsonian}) that measure the alignment between an image and a text prompt. Specifically, we show that a simple metric, {\bf VQAScore}~\cite{lin2024evaluating}, which computes the likelihood of generating a ``Yes'' answer to a question like ``Does this figure show \{text\}?'' from a VQA model, significantly surpasses previous metrics in correlating with human judgments. VQAScore can be calculated end-to-end from off-the-shelf VQA models, without finetuning on human feedback~\cite{pickscore, imagereward} or decomposing prompts into QA pairs~\cite{vq2, davidsonian}. VQAScore is strong because it leverages the compositional reasoning capabilities of recent multimodal large language models (LLMs)~\cite{llava15, instructblip} trained for VQA. For instance, our study adopts the open-source {\bf CLIP-FlanT5} model~\cite{lin2024evaluating}, which uses a bidirectional encoder that allows the image and question embeddings to ``see'' each other. VQAScore based on CLIP-FlanT5 sets a new state-of-the-art on both GenAI-Bench and previous benchmarks like TIFA160~\cite{tifa} and Winoground~\cite{winoground}. As such, we recommend VQAScore over the ``bag-of-words'' CLIPScore, which has been widely misused in our community~\cite{aro, kamath2023text}. We will release all human ratings to facilitate the development of automated metrics.

{\bf Improving generation with VQAScore.} We show that text-to-image generation can be improved by selecting the candidate image with the highest VQAScore (from a set of candidates). This ranking-based approach does not require any finetuning and can operate in a fully black-box manner~\cite{liu2024language}, needing only an image generation API. Notably, simply ranking between 3 to 9 images can already enhance the average human ratings for DALL-E 3 and SD-XL by 0.2 to 0.3 (on a 1-to-5 Likert scale), setting the new closed-source and open-source SOTAs on GenAI-Bench. VQAScore significantly outperforms other metrics; for instance, using CLIPScore for ranking often leads to the same or lower human ratings. We present qualitative examples in \autoref{fig:ranking_dalle3}. Overall, VQAScore emerges as the most effective ranking metric, surpassing other metrics that rely on costly human feedback (e.g., PickScore~\cite{pickscore}) or ChatGPT for prompt decomposition (e.g., Davidsonian~\cite{davidsonian}) by 2x to 3x. 

{\bf Limitations.} Lastly, we explore the implications of Goodhart's Law~\cite{goodhart2015goodhart}, particularly limitations of VQAScore in detecting fine-grained visual details and resolving linguistic ambiguity. Despite these mild limitations, we strongly urge the research community to adopt VQAScore as a reproducible supplement to non-reproducible human studies~\cite{otani2023toward}, or as a more reliable alternative to CLIPScore, which has ceased to be effective~\cite{kamath2023text, aro, lin2024revisiting}.

\begin{table*}[h!]
\centering
\caption{{\bf Comparing GenAI-Bench to existing text-to-visual benchmarks.} GenAI-Bench covers more essential skills of compositional text-to-visual generation, emphasizing advanced reasoning skills (highlight in \textcolor{myblue}{blue}) that are required to parse complex user prompts. Moreover, GenAI-Bench tags each prompt with all its evaluated skills, whereas most benchmarks assign no tags or only one or two per prompt (even when multiple skills are involved). GenAI-Bench also provides human ratings for both image and video generative models to benchmark automated metrics.}
\scalebox{0.7}{
    \begin{tabular}{@{}lcccccccccc@{}}
    \toprule
    \multirow{2}{*}{\bf Benchmarks} & \multicolumn{8}{c}{\bf  Skills Covered in Compositional Text-to-Visual Generation} & \multirow{2}{*}{\bf Tagging} & \multirow{2}{*}{\bf Human Annotation} \\ \cmidrule{2-9}
    \cmidrule(l){2-4} \cmidrule(l){5-9}
     & {\footnotesize \textcolor{gray}{Scene}} & {\footnotesize \textcolor{gray}{Attribute}} & {\footnotesize \textcolor{gray}{Relation}} & {\footnotesize \textcolor{myblue}{Count}} & {\footnotesize \textcolor{myblue}{Negation}} & {\footnotesize \textcolor{myblue}{Universal}} & {\footnotesize \textcolor{myblue}{Compare}} & {\footnotesize \textcolor{myblue}{Differ}}\\ 
    \midrule
    PartiPrompt (P2)~\cite{parti} & \cmark & \cmark & \cmark & \cmark  & \cmark & \xmark & \xmark & \xmark & 2 Tags & \xmark \\ 
    DrawBench~\cite{imagen} & \cmark & \cmark & \cmark & \cmark & \xmark & \xmark & \xmark & \xmark & 1 Tag & \xmark\\ 
    EditBench~\cite{wang2023imagen} & \cmark & \cmark & \cmark & \cmark & \xmark & \xmark & \xmark & \xmark & \xmark & \xmark \\ 
    TIFAv1~\cite{tifa} & \cmark & \cmark & \cmark & \cmark  & \xmark & \xmark & \xmark & \xmark & All Tags & Images \\ 
    Pick-a-pic~\cite{pickscore} & \cmark & \cmark & \cmark & \cmark & \xmark& \xmark & \xmark & \xmark & \xmark & Images \\ 
    T2I-CompBench~\cite{huang2023t2i} & \cmark & \cmark & \cmark & \cmark & \xmark & \xmark & \xmark & \xmark & 1 Tag & Images \\ 
    HPDv2~\cite{hpsv2} & \cmark & \cmark & \cmark & \xmark & \xmark & \xmark & \xmark & \xmark & \xmark & Images \\ 
    EvalCrafter~\cite{evalcrafter} & \cmark & \cmark & \cmark & \cmark & \xmark & \xmark & \xmark & \xmark & \xmark & Videos \\ 
    \midrule 
    {\bf GenAI-Bench (Ours)} & \cmark & \cmark & \cmark & \cmark & \cmark & \cmark & \cmark  & \cmark & All Tags & Images \& Videos \\ 
    \bottomrule
    \end{tabular}
}
\label{tab:skill_comparison}
\end{table*}

{\bf Contribution summary.}
\begin{enumerate}
    \item We conduct an extensive human study on compositional text-to-visual generation using {\bf GenAI-Bench}, revealing limitations of leading open-source and closed-source models. 
    \item We present a simple black-box approach that improves generation by {\bf ranking} images with VQAScore, significantly surpassing other scoring methods by 2x to 3x. 
    \item We will release {\bf GenAI-Rank} with over 40,000 human ratings to benchmark methods that rank images generated from the same prompt.
\end{enumerate}

\section{Related Works}

{\bf Text-to-visual benchmarks.} Early benchmarks mostly rely on captions from existing datasets like COCO~\cite{coco, dalle2, dalleval, tifa}, focusing on generating simple objects, attributes, and scenes. Other benchmarks, such as HPDv2~\cite{hpsv2} and Pick-a-pic~\cite{pickscore}, primarily evaluate image quality (aesthetic) using simpler text prompts. Recently, benchmarks like DrawBench~\cite{imagen}, PartiPrompt~\cite{parti}, and T2I-CompBench~\cite{huang2023t2i} have shifted the focus to compositional text-to-image generation with an emphasis on attribute bindings and object relationships. Our GenAI-Bench escalates the challenge by incorporating more practical prompts that require ``advanced'' reasoning (e.g., logic and comparison) to benchmark next-generation text-to-visual models.

{\bf Automated metrics.} Perceptual metrics like IS~\cite{IS}, FID~\cite{FID} and LPIPS~\cite{LPIPS} use pre-trained networks to assess the quality of generated imagery using reference images. To evaluate vision-language alignment (also referred to as faithfulness or consistency~\cite{tifa, dai2023emu, manas2024improving}), recent studies~\cite{chang2023muse, wu2023tune, singer2022make, ruiz2023dreambooth, kawar2023imagic, gal2022image, kumari2023multi, meng2021sdedit, gal2022stylegan} primarily report CLIPScore~\cite{clipscore}, which measures (cosine) similarity of the embedded image and text prompt. However, CLIP cannot reliably process compositional text prompts due to its ``bag-of-words'' encoding~\cite{kamath2023text, aro, lin2024revisiting}. Human preference models like ImageReward~\cite{imagereward}, PickScore~\cite{pickscore}, and HPSv2~\cite{hpsv2} further leverage human feedback to improve models like CLIP by finetuning on large-scale human ratings. Another popular line of works~\cite{tifa, huang2023t2i, vpeval, singh2023divide, t2vscore} uses LLMs like ChatGPT to decompose texts into simpler components for analysis, e.g., via question generation and answering (QG/A)~\cite{davidsonian}. For example, Davidsonian Scene Graph~\cite{davidsonian} decomposes a text prompt into simpler QA pairs and outputs a score as the accuracy of answers generated by a VQA model. However, Lin et al.~\cite{lin2024revisiting, lin2024evaluating} show that such methods still struggle with complex text prompts and propose VQAScore, an end-to-end metric that better correlates with human judgments.

\section{GenAI-Bench for Text-to-Visual Evaluation}
\label{sec:genai_bench}
In this section, we review {\bf GenAI-Bench}~\cite{lin2024evaluating}, a challenging benchmark featuring real-world user prompts that cover essential aspects of compositional text-to-visual generation. %

{\bf Skill taxonomy.} Prior literature on text-to-visual generation~\cite{parti, imagen, huang2023t2i} focuses on generating ``basic'' objects, attributes, relations, and scenes. However, as shown in \autoref{fig:teaser}, user prompts often need ``advanced'' compositional reasoning, including comparison, differentiation, counting, and logic. These ``advanced'' compositions extend beyond the ``basic'' ones. For example, user prompts may involve counting not just objects, but also complex object-attribute-relation compositions, e.g., ``{\tt three white seagulls flying over a blue lake}''. Also, users often want to generate multiple objects of the same category but differentiate them with varied properties, e.g., ``{\tt three flowers on the ground; one red, another yellow, and the third blue}''. To this end, we collaborate with designers who use text-to-image tools~\cite{midjourney} in their profession to design a taxonomy that includes both ``basic'' (objects, scenes, attributes, and spatial/action/part relations) and ``advanced'' skills (counting, comparison, differentiation, negation, and universality). \autoref{tab:skill_comparison} shows that GenAI-Bench uniquely covers all these essential skills. \appref{sec:skill_details} provides definitions and more examples.

{\bf GenAI-Bench.} We also collect 1,600 prompts from these professional designers. Importantly, collecting prompts from actual designers ensures that they are free from subjective (non-visual) or toxic contents. For example, we observe that ChatGPT-generated prompts from T2I-CompBench~\cite{huang2023t2i} can include subjective phrases like ``{\tt a natural symbol of rebirth and renewal}''. We also find that the Pick-a-pic~\cite{pickscore} contains inappropriate prompts (e.g., those suggesting NSFW images of celebrities) created by malicious web users. To avoid legal issues and test general model capabilities, we instruct the designers to create prompts with generic characters and subjects (e.g., food, vehicles, humans, pets, plants, household items, and mythical creatures) instead of celebrities or copyrighted characters. \appref{sec:genai_bench_details} details our collection procedure and discuss how we avoid these issues. For a fine-grained analysis, we tag each prompt with {\em all} its evaluated skills. This contrasts with previous benchmarks that provide no tags~\cite{hpsv2, pickscore, evalcrafter} or only assign only one or two tags per prompt~\cite{parti, imagen, huang2023t2i}. In total, GenAI-Bench provides over 5,000 human-verified tags with an approximately balanced distribution of skills: about half of the prompts involve only ``basic'' compositions, while the other half poses greater challenges by incorporating both ``basic'' and ''advanced'' compositions.
 
\section{Human Evaluation via GenAI-Bench}
\label{sec:human_eval}
We now present an extended human study of ten popular image and video generative models.

{\bf Human evaluation.} We evaluate six text-to-image models: Stable Diffusion~\cite{stablediffusion} (SD v2.1, SD-XL, SD-XL Turbo), DeepFloyd-IF~\cite{deepfloyd}, Midjourney v6~\cite{midjourney}, DALL-E 3~\cite{dalle3}; along with four text-to-video models: ModelScope~\cite{modelscope}, Floor33~\cite{floor33}, Pika v1~\cite{pikalab}, Gen2~\cite{gen2}. Next, we hire three annotators to rate on a 1-to-5 Likert scale for image-text or video-text alignment using the recommended annotation protocol of \cite{otani2023toward}:
\begin{mdframed}[linewidth=1pt, linecolor=black, leftmargin=3.7cm, rightmargin=3.7cm, backgroundcolor=gray!20, roundcorner=5pt]
\scriptsize
\textcolor{blue}{How well does the image (or video) match the description?} \\
\quad 1. Does not match at all.\\
\quad 2. Has significant discrepancies.\\
\quad 3. Has several minor discrepancies.\\
\quad 4. Has a few minor discrepancies.\\
\quad 5. Matches exactly.
\end{mdframed}

\begin{figure}[h!]
\centering
    \scalebox{0.94}{
        \begin{tabular}{cc}
           (a) GenAI-Bench (Image Ratings) &  (b) GenAI-Bench (Video Ratings)  \\
        \includegraphics[height=0.10\textheight]{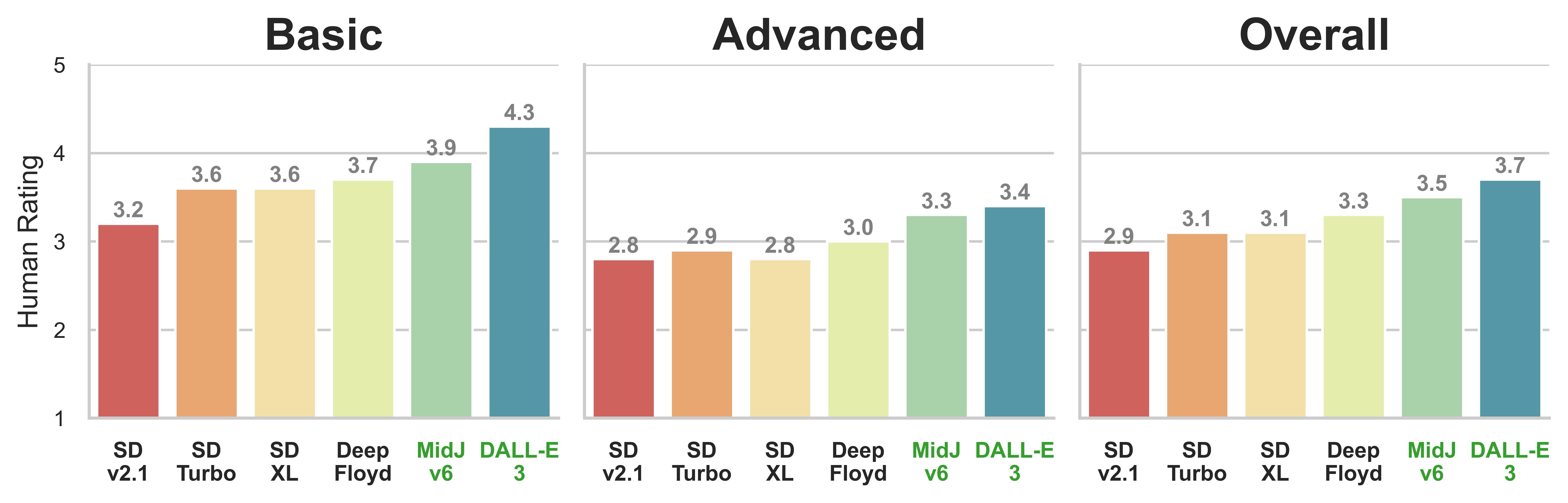}  & \includegraphics[height=0.10\textheight]{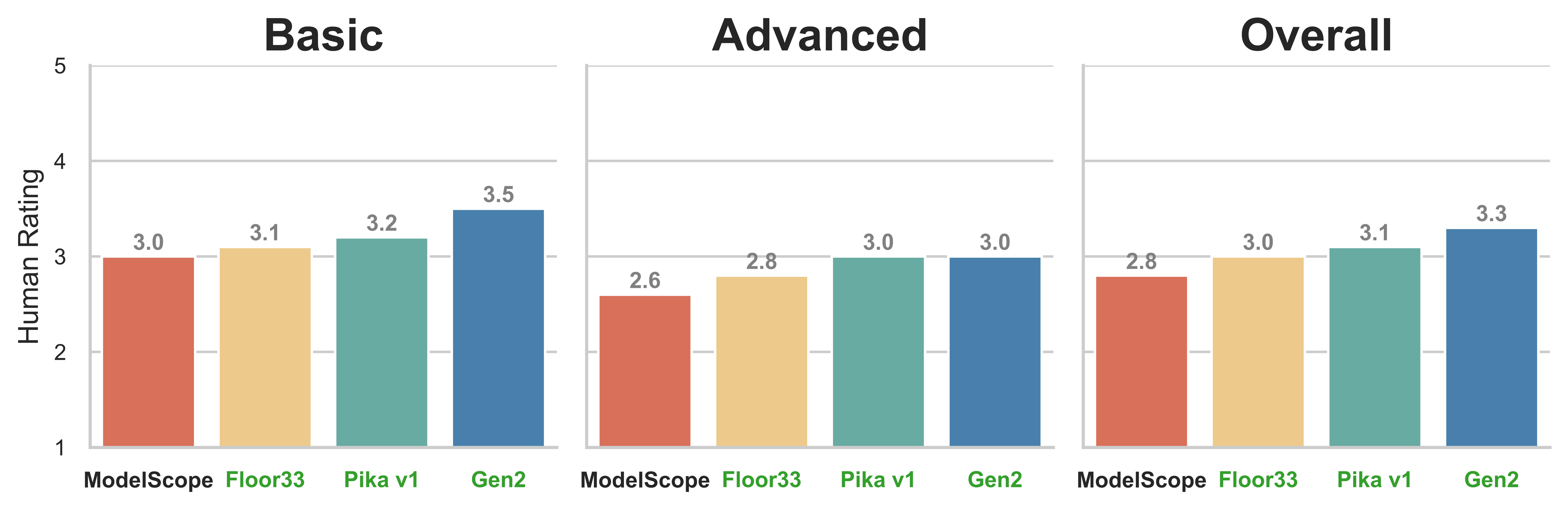}
        \end{tabular}
    }
    \caption{\small {\bf Human evaluation on GenAI-Bench.} We show the average human alignment ratings on ten popular image and video generative models. We highlight closed-source models (e.g., DALL-E 3~\cite{dalle3}) in \textcolor{ForestGreen}{green}. We find that (1) ``advanced'' prompts that require higher-order reasoning (e.g., negation and comparison) challenge all models more, (2) models using better text embeddings or captions (DeepFloyd-IF~\cite{deepfloyd} and DALL-E 3~\cite{dalle3}) outperform others (SD-XL~\cite{stablediffusion}), (3) open-source and video-generative models~\cite{stablediffusion, gen2} still lag behind their closed-sourced and image-generative counterparts, suggesting room for improvement.}
    \label{fig:genai_bench_leaderboard}
\end{figure}

\begin{figure*}[h!]
\centering
    \scalebox{1.0}{\includegraphics[height=0.19\textheight]{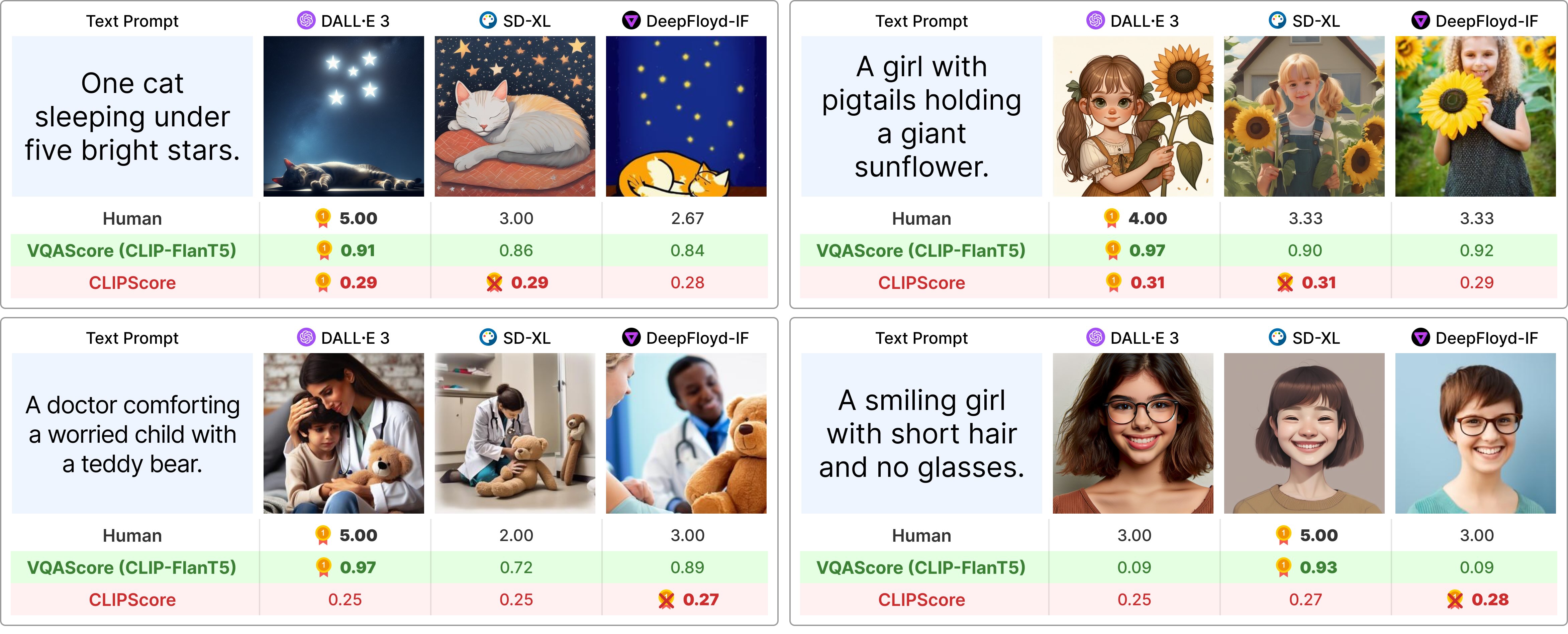} 
    }
    \caption{\small {\bf VQAScore (based on CLIP-FlanT5~\cite{lin2024evaluating}) versus CLIPScore} on samples from GenAI-Bench. VQAScore shows a significantly stronger agreement with human ratings compared to CLIPScore~\cite{clipscore}, making it a more reliable tool for automatic text-to-visual evaluation, especially on user prompts that involve complex compositional reasoning.}
    \label{fig:vqascore_teaser_1}
\end{figure*}

\noindent Our collected human ratings indicate a high level of inter-rater agreement, with Krippendorff's Alpha reaching 0.72 for image ratings and 0.70 for video ratings, suggesting substantial agreement~\cite{tifa}. The use of the Likert scale also makes the final rating interpretable. For example, a score near 5 implies that the model's generated images almost always ``{\em match exactly}'' with the input prompts.

{\bf Analysis.} \autoref{fig:genai_bench_leaderboard} presents human ratings for basic, advanced, and overall prompts. Notably, advanced prompts that require complex visio-linguistic reasoning are much harder. For example, the top-performing DALL-E 3 scores $4.3$ on basic prompts, indicating ``{\em a few minor discrepancies}''. However, on advanced prompts, its score drops to $3.4$, indicating ``{\em several minor discrepancies}''. Interestingly, models (e.g., DeepFloyd-IF and DALL-E 3) using stronger text embeddings from LLMs (e.g., T5~\cite{t5}) outperform those using CLIP text embeddings (e.g., SD-XL). Lastly, we observe that open-source and video-generative models lag behind their closed-source and image-generative counterparts, suggesting room for future innovation. In \appref{sec:genai_bench_details}, we detail model performance across various skills, highlighting challenges in higher-order reasoning like negation and comparison.

\section{Evaluating Automated Metrics}
\label{sec:metrics}
We use our human ratings to benchmark automated alignment metrics~\cite{clipscore, vq2, davidsonian} on GenAI-Bench. 

\begin{table}[h!]
\centering
\renewcommand{\arraystretch}{1.3}
\caption{{\bf Evaluating the correlation of automated metrics with human ratings on GenAI-Bench.} We report Pairwise accuracy~\cite{deutsch2023ties}, Pearson, and Kendall, with higher scores indicating better performance for all. {\bf VQAScore} based on the CLIP-FlanT5 model~\cite{lin2024evaluating} achieves the strongest agreement with human ratings on images and videos, significantly surpassing popular metrics like CLIPScore~\cite{clipscore}, PickScore~\cite{pickscore}, and Davidsonian~\cite{davidsonian}.
}
\scalebox{0.65}{
\begin{tabular}{c}
\scalebox{1.0}{
\begin{NiceTabular}{lcccccc}
    \CodeBefore
    \Body
\toprule[1.2pt]
\multirow{2}{*}{\textbf{Method}} & \multicolumn{3}{c}{{\bf GenAI-Bench (Image)}}  & \multicolumn{3}{c}{{\bf GenAI-Bench (Video)}} \\ \cmidrule(r){2-4} \cmidrule(l){5-7}
& {\textbf{\small Pairwise}} & \textbf{\small Pearson} & \textbf{\small Kendall} & {\textbf{\small Pairwise}} & \textbf{\small Pearson} & \textbf{\small Kendall} \\
\midrule
CLIPScore~\cite{clipscore} & 50.8 & 16.4 & 11.8 & 53.6 & 25.3 & 18.0 \\
BLIPv2Score~\cite{clipscore} & 52.2 & 17.2 & 14.7 & 54.6 & 25.3 & 20.1 \\
ImageReward~\cite{imagereward} & 56.6 & 35.0 & 24.0 & 60.0 & 42.9 & 31.4  \\ 
PickScore~\cite{pickscore} & 57.1  & 35.4 & 25.0 & 56.8  & 34.6 & 24.8  \\ 
HPSv2~\cite{hpsv2} & 49.6 & 13.9 & 9.6 & 51.5 & 18.4 & 13.7 \\
LLMScore~\cite{llmscore} & 53.2 & 15.4 & 13.6 & 53.2 & 19.4 & 17.7 \\
BLIP-VQA~\cite{huang2023t2i} & 54.3 & 27.1 & 23.0 & 55.1 & 29.8 & 22.5 \\
VQ2~\cite{vq2} & 51.9 & 13.3 & 12.0 & 52.8 & 18.0 & 15.5 \\
Davidsonian~\cite{davidsonian} & 54.6 & 29.3 & 22.4 & 55.9 & 32.3 & 23.5 \\
\midrule
{\bf VQAScore~\cite{lin2024evaluating}} & {\bf 64.1} & {\bf 49.9} & {\bf 39.8} & {\bf 63.2} & {\bf 50.6} & {\bf 38.2}  \\
\bottomrule[1.2pt]
\end{NiceTabular}
}
\end{tabular}
}
\label{tab:genai_bench_human_correlation}
\end{table}

{\bf VQAScore.} Given an {\tt image} and {\tt text}, we calculate the probability of a ``Yes'' answer to a simple {\tt question} like ``{Does this figure show `\{{\tt text}\}'? Please answer yes or no.}'':
\begin{equation}
    P(\text{``Yes''} | \text{\tt image}, \text{\tt question})
\label{eq:vqa}
\end{equation}
\noindent We implement VQAScore using the open-source\footnote{One can now run VQAScore using proprietary models such as GPT-4o and Gemini-1.5 using their log-prob features. Please refer to our codebase at \href{https://github.com/linzhiqiu/t2v_metrics}{link}.} {\bf CLIP-FlanT5} model~\cite{lin2024evaluating} trained on 665K public VQA data~\cite{llava15}. For video-text pairs, we average the scores across all video frames following~\cite{singer2022make}. 

{\bf Evaluation setup.} To evaluate automated metrics on GenAI-Bench, we follow \cite{tifa} to report the Pearson and Kendall coefficients, which measures the correlation of the metric score with human judgment. However, \cite{deutsch2023ties} (EMNLP'23 outstanding paper) show several issues with these metrics. For example, Pearson assumes a linear relationship between metric and human scores, while Kendall skips ties common in 1-to-5 Likert scales. As such, we report {\bf Pairwise} accuracy~\cite{deutsch2023ties}, which is designed to address these issues. We refer readers to \cite{deutsch2023ties} for detailed equations and provide an overview below. For a dataset with $M$ items (e.g., image-text pairs), there are two $M$-size score vectors: one for human ratings and one for metric scores. Pairwise accuracy (a value between 0 and 1) evaluates the percentage of agreement across all $M$ x $M$ pairs of items, that is, if one item scores higher, lower, or ties with another item in both human and metric scores. Lastly, we apply the tie calibration technique from \cite{deutsch2023ties} to find the optimal tie threshold for each metric.

{\bf Results.} \autoref{tab:genai_bench_human_correlation} shows that VQAScore significantly outperforms previous metrics such as CLIPScore~\cite{clipscore}, models trained with extensive human feedback~\cite{imagereward,hpsv2,pickscore}, and QG/A methods that use the same CLIP-FlanT5 VQA model~\cite{vq2,davidsonian}. In addition, we implement BLIP-VQA~\cite{huang2023t2i} and LLMScore~\cite{llmscore} using their official codebase. \appref{sec:genai_bench_details} shows that VQAScore achieves the best performance across all ``basic'' and ``advanced'' skills on GenAI-Bench. \appref{sec:method_details} shows that VQAScore also achieves the state-of-the-art performance on seven more alignment benchmarks such as TIFA160~\cite{tifa} and Winoground~\cite{winoground}. \autoref{fig:vqascore_teaser_1} qualitatively compare VQAScore against CLIPScore on random samples from GenAI-Bench. The strong performance of VQAScore makes it a more reliable tool for the future automated evaluation of text-to-visual models.

\section{Improving Text-to-Visual Generation}
\label{sec:improving}
VQAScore's superior performance in evaluating text-to-visual generation suggests its potential to improve generation as well. We now show that VQAScore can improve the alignment of DALL-E 3~\cite{dalle3} and SD-XL~\cite{stablediffusion} by simply ranking candidate images. We also collect a {\bf GenAI-Rank} benchmark to evaluate scoring metrics on ranking images from the same prompt.

\begin{figure}[h!]
\centering
    \scalebox{0.5}{
        \begin{tabular}{c}
           \includegraphics[width=0.96\columnwidth]{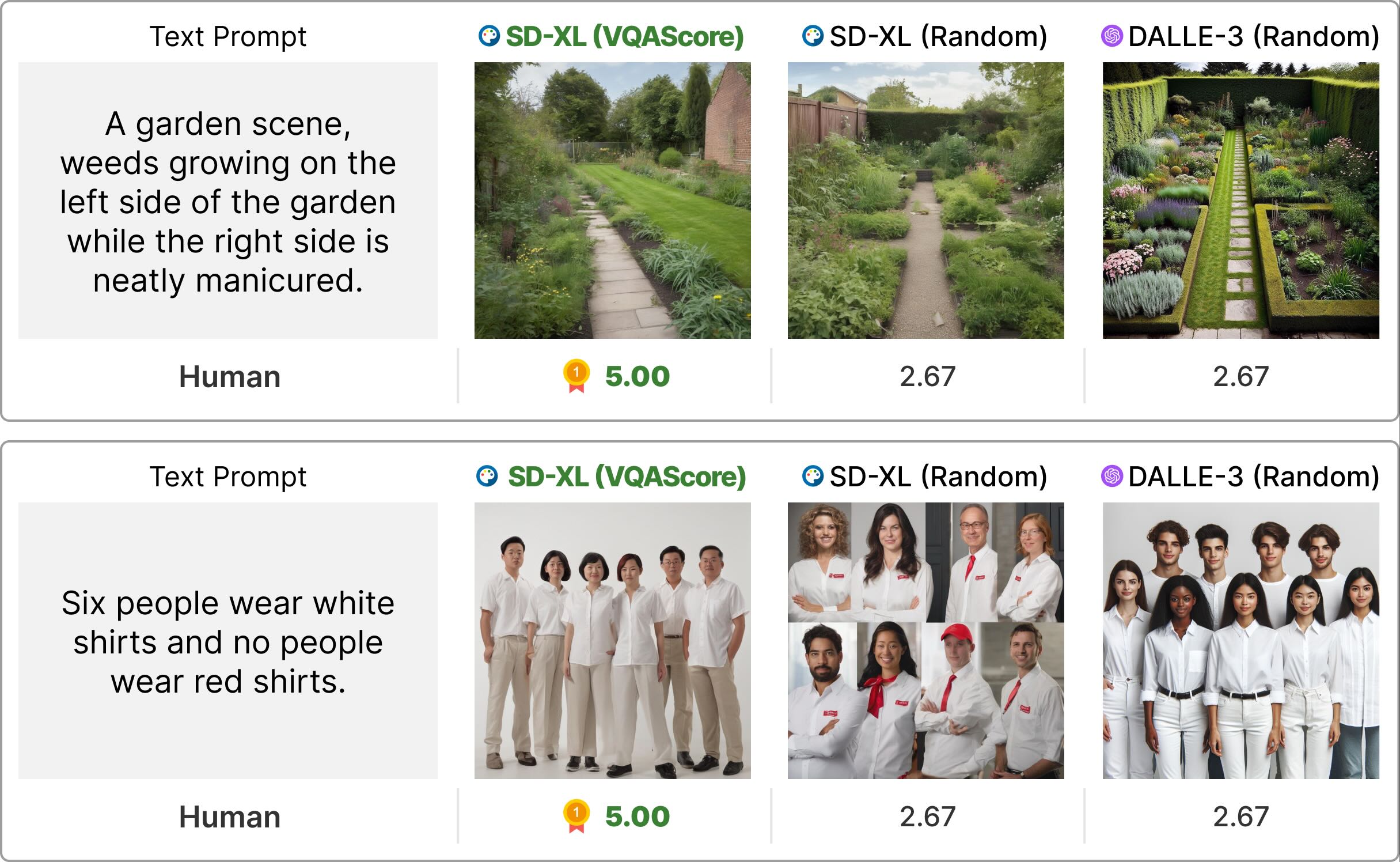} 
        \end{tabular}
    }
    \caption{\small {\bf VQAScore can select images generated by SD-XL~\cite{stablediffusion} that outperform DALL-E 3's~\cite{dalle3}.} Although less powerful in prompt alignment than DALL-E 3, SD-XL~\cite{stablediffusion} can still be improved by selecting the highest VQAScore image from merely three candidates. We provide examples of how VQAScore ranks SD-XL images in \appref{sec:additional_examples}. }
    \label{fig:sd_beats_dalle3}
\end{figure}

\begin{figure}[h!]
\centering
    \scalebox{0.65}{
        \begin{tabular}{cc}
           {\large (a) Improving DALL-E 3 by image ranking} & {\large (b) Improving SD-XL by image ranking} \\
        \includegraphics[height=0.2\textheight]{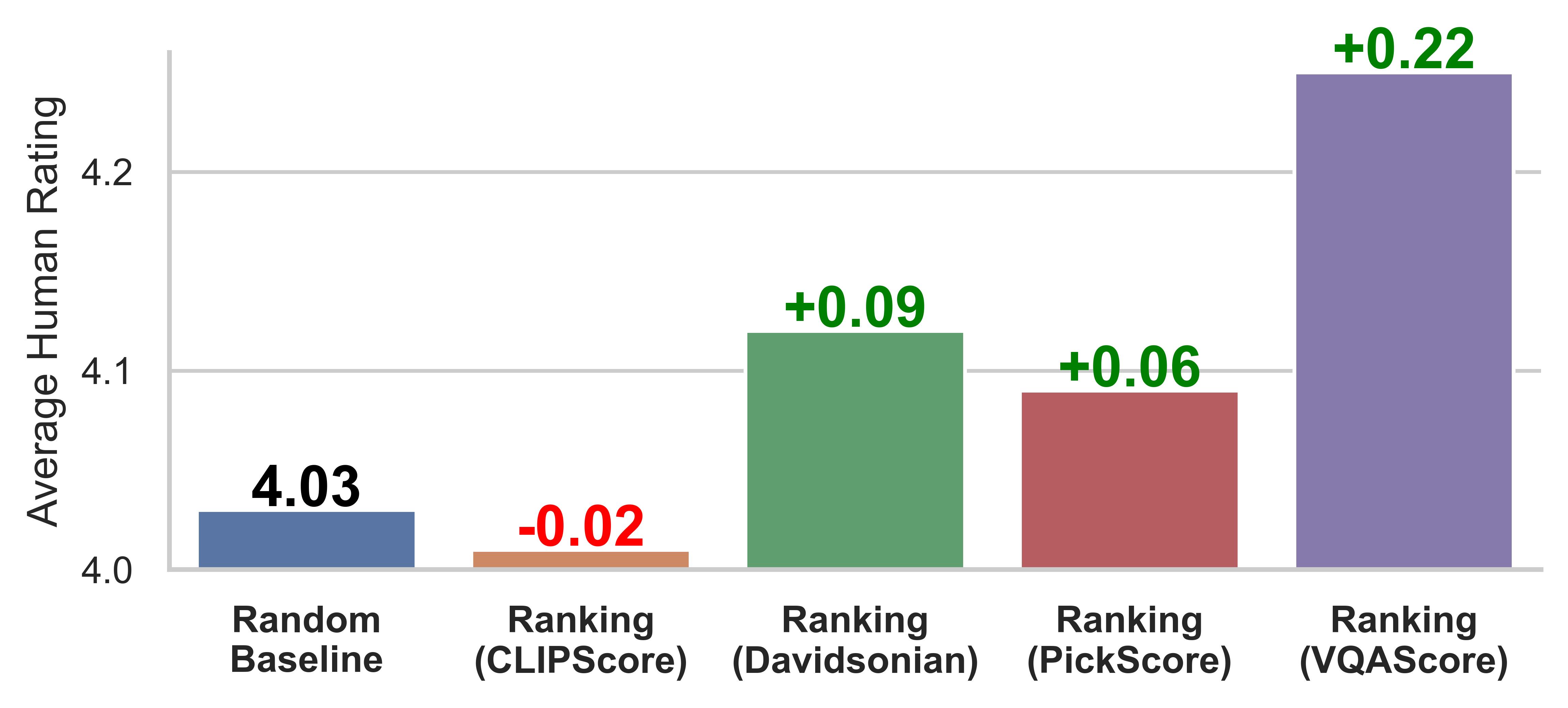} &
        \includegraphics[height=0.2\textheight]{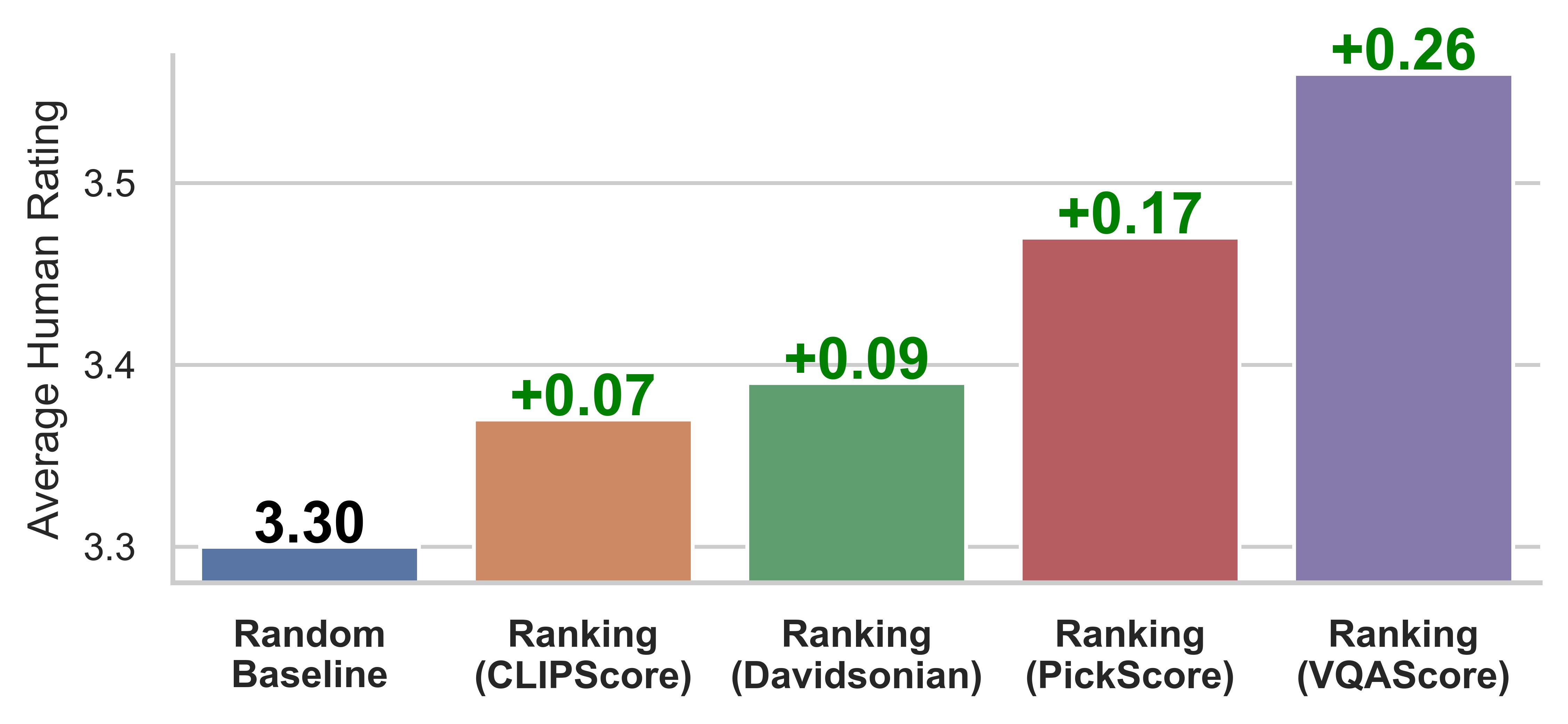} \\
        
        \end{tabular}
    }
    \caption{\small {\bf Improving text-to-visual generation by ranking nine candidate images.} We show the performance gains over the {\em Random} baseline (no ranking) in \textcolor{ForestGreen}{green} and decreases in \textcolor{red}{red}. Notably, selecting the highest-VQAScore images from nine candidates significantly boosts the overall human alignment ratings. In contrast, ranking by CLIPScore~\cite{clipscore} results in the same or lower performance. Overall, VQAScore is 2x to 3x more effective than other methods that rely on costly human feedback (PickScore~\cite{pickscore}) or decompose texts using ChatGPT (Davidsonian~\cite{davidsonian}). \autoref{tab:ranking_all} reports more scoring methods.}
    \label{fig:ranking_all}
\end{figure}

\begin{figure*}[htbp!]
\centering
    \scalebox{0.9}{
        \begin{tabular}{c@{\hspace{1pt}}c}
           \includegraphics[width=0.5\textwidth]{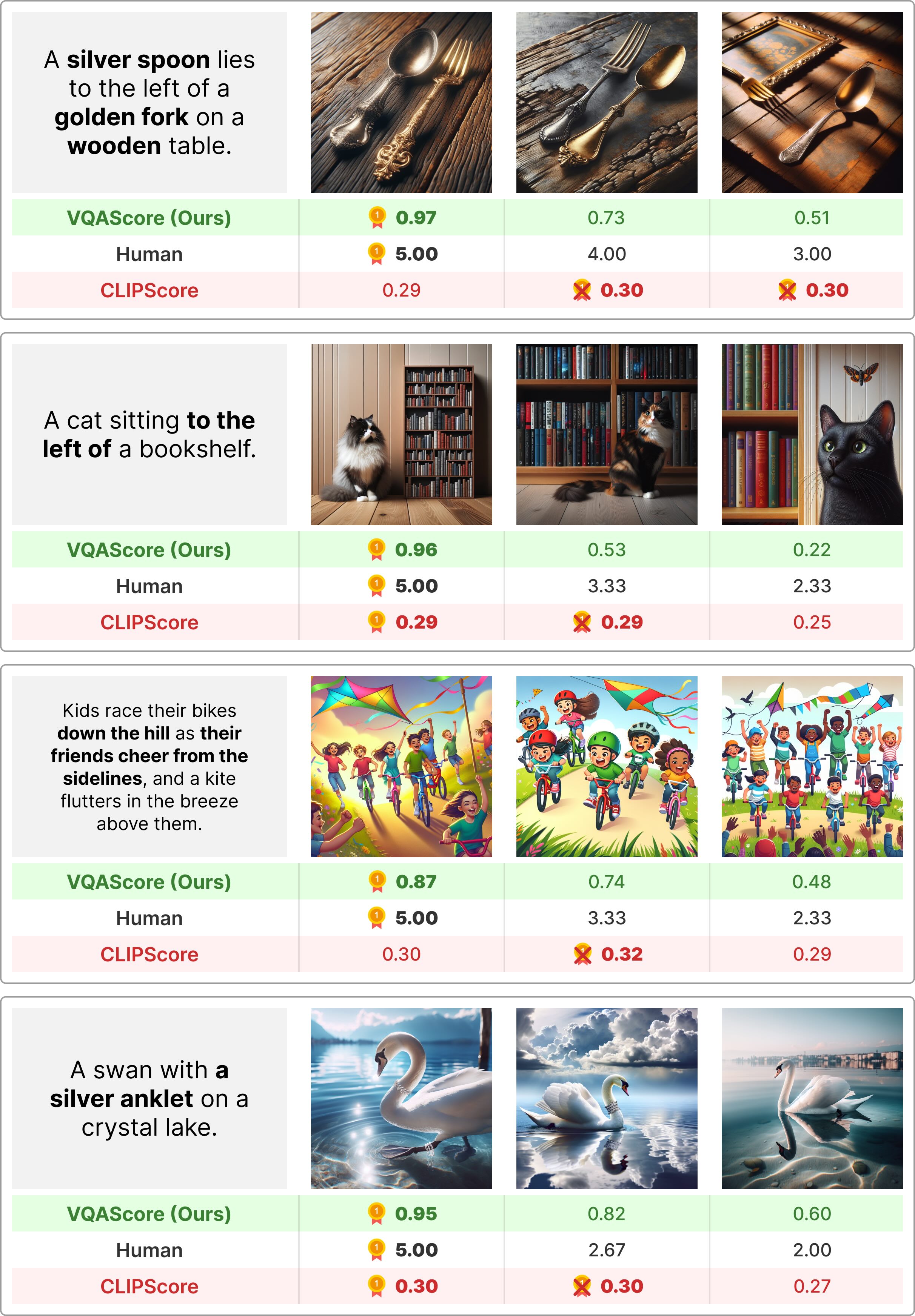} & \includegraphics[width=0.5\textwidth]{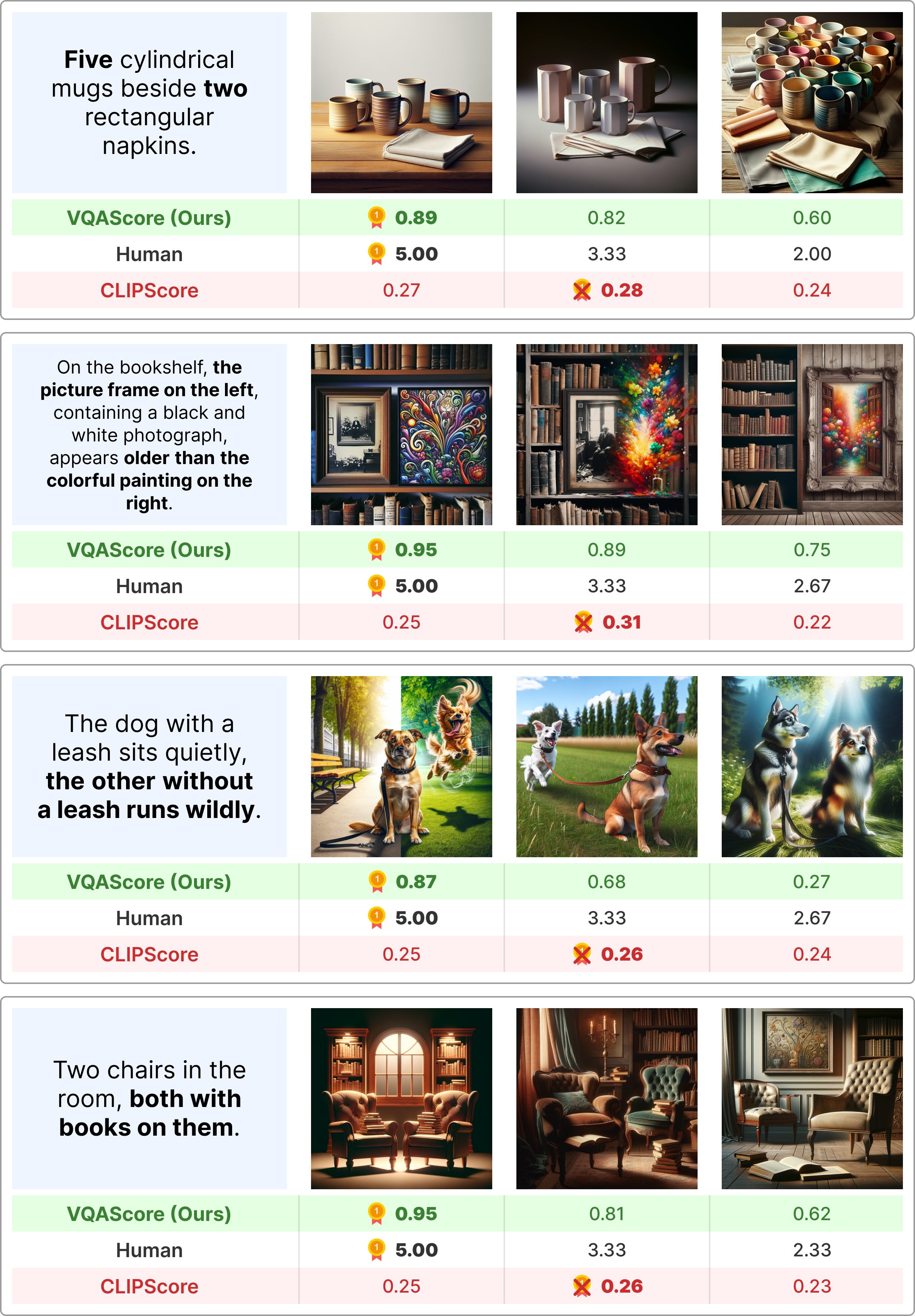} \\
        \end{tabular}
    }
    \caption{\small {\bf Ranking DALL-E 3 generated images with VQAScore or CLIPScore.} VQAScore outperforms CLIPScore in ranking candidate images generated by DALL-E 3, particularly for prompts that involve attributes, relationships, and higher-order reasoning. This indicates that VQAScore can already improve text-to-image generation using only an image generation API~\cite{liu2024language}. 
    }
    \label{fig:ranking_dalle3}
\end{figure*}

{\bf Ranking images by VQAScore.} Given the same prompt, most text-to-visual models produce vastly different images with each run. As such, we adopt a {\em black-box} method~\cite{liu2024language} that improves text-to-image generation by selecting the highest-VQAScore image from a few generated candidates. This ranking-based approach is simple yet surprisingly effective. For instance, despite SD-XL's weaker prompt alignment compared to DALL-E 3, \autoref{fig:sd_beats_dalle3} shows how VQAScore can select the best SD-XL images (from three candidates) that outperform DALL-E 3's. \autoref{fig:ranking_dalle3} shows that VQAScore can also improve the closed-source (black-box) DALL-E 3 by correctly selecting the most prompt-aligned images from three candidates.

{\bf GenAI-Rank: A benchmark for text-to-image ranking.} To compare against other ranking metrics (e.g., CLIPScore and PickScore), we hire three annotators to rate nine generated images for each prompt. In this study, we randomly select 800 prompts from GenAI-Bench and collect 43,200 human ratings for 14,400 images generated by DALL-E 3 and SD-XL. We will release this benchmark (termed {\bf GenAI-Rank}) for reproducibility and to facilitate the evaluation of future ranking metrics.

{\bf VQAScore achieves superior performance gains.} \autoref{fig:ranking_all} confirms that ranking by VQAScore delivers the most significant improvements in human ratings.  While ranking by CLIPScore~\cite{clipscore} results in the same or even lower performance, VQAScore consistently improves with more images to rank. VQAScore is also 2x to 3x more effective than other ranking metrics that rely on expensive human feedback (e.g., PickScore~\cite{pickscore}) or decompose texts via ChatGPT (e.g., Davidsonian~\cite{davidsonian}). \autoref{tab:ranking_all} details the performance gains for ranking 3 to 9 images across basic, advanced, and all prompts. VQAScore notably improves the prompt alignment of DALL-E 3 and SD-XL by about $0.3$ on ``advanced'' prompts that require complex visio-linguistic reasoning, such as counting, comparison, and logic. Lastly, although we use the open-source CLIP-FlanT5 model for this study, future work may use stronger models such as GPT-4o for improved performance.

\begin{table*}[htbp!]
\centering
\renewcommand{\arraystretch}{1.3}
\caption{{\bf Comparing scoring methods for image ranking on GenAI-Rank.} We present the average human ratings of 7 popular scoring methods across basic, advanced, and all prompts. Performance gains over the {\em Random} baseline (no ranking) are highlighted in \textcolor{ForestGreen}{green}, while decreases are marked in \textcolor{red}{red}. We first note that ranking by CLIPScore~\cite{clipscore} can lead to a performance drop. For instance, CLIPScore results in a $0.04$ drop when given more images to rank (from 3 to 9). In contrast, VQAScore demonstrates consistent and significant improvements with more images. VQAScore improves performance on the more challenging ``advanced'' prompts that require complex visio-linguistic reasoning skills like counting, comparison, and logic. For these ``advanced'' prompts, VQAScore boosts DALL-E 3 by $0.30$ and SD-XL by $0.27$ by ranking nine images, outperforming the second-best method PickScore~\cite{pickscore} by 2x to 3x. For reference, we include human (oracle) performance (ranking by ground-truth human ratings). GenAI-Rank releases all human ratings to help benchmark future ranking metrics.
}
\scalebox{0.5}{
\begin{tabular}{cc}
\scalebox{1.0}{
\begin{NiceTabular}{lcccccc}
    \CodeBefore
      \rectanglecolor{softgray}{3-1}{4-7}
      \rectanglecolor{softgreen}{11-1}{11-7}
    \Body
\toprule[1.2pt]
\multirow{2}{*}{\textbf{Method}} & \multicolumn{2}{c}{{\bf Basic}}  & \multicolumn{2}{c}{{\bf Advanced}} & \multicolumn{2}{c}{{\bf Overall}}\\ \cmidrule(r){2-3} \cmidrule{4-5}\cmidrule(l){6-7}
& {\textbf{\small 3 Imgs}} & \textbf{\small 9 Imgs} & \textbf{\small 3 Imgs} & {\textbf{\small 9 Imgs}} & \textbf{\small 3 Imgs} & \textbf{\small 9 Imgs} \\
\midrule
Random & $4.51 $ & $4.51 $ & $3.77$ & $3.77$ & $4.03$& $4.03$\\
Human Oracle & {\bf $4.77_{\textcolor{ForestGreen}{+.26}}$} 
&{\bf $4.89_{\textcolor{ForestGreen}{+.38}}$}
& {\bf $4.18_{\textcolor{ForestGreen}{+.41}}$}
& {\bf $4.46_{\textcolor{ForestGreen}{+.69}}$}
& {\bf $4.39_{\textcolor{ForestGreen}{+.36}}$}
& {\bf $4.61_{\textcolor{ForestGreen}{+.58}}$} \\
\midrule
CLIPScore~\cite{clipscore}  & {\bf $4.54_{\textcolor{ForestGreen}{+.03}}$} 
& {\bf $4.53_{\textcolor{ForestGreen}{+.02}}$}
& {\bf $3.79_{\textcolor{ForestGreen}{+.02}}$}
& {\bf $3.73_{\textcolor{red}{-.04}}$}
& {\bf $4.05_{\textcolor{ForestGreen}{+.02}}$}
& {\bf $4.01_{\textcolor{red}{-.02}}$}  \\

ImageReward~\cite{imagereward}  & {\bf $4.56_{\textcolor{ForestGreen}{+.05}}$}
& {\bf $4.52_{\textcolor{ForestGreen}{+.01}}$}
& {\bf $3.82_{\textcolor{ForestGreen}{+.05}}$}
& {\bf $3.83_{\textcolor{ForestGreen}{+.06}}$}
& {\bf $4.08_{\textcolor{ForestGreen}{+.05}}$}
& {\bf $4.08_{\textcolor{ForestGreen}{+.05}}$}  \\

PickScore~\cite{pickscore}  & {\bf $4.58_{\textcolor{ForestGreen}{+.07}}$} 
& {\bf $4.60_{\textcolor{ForestGreen}{+.09}}$}
& {\bf $3.82_{\textcolor{ForestGreen}{+.05}}$} 
& {\bf $3.81_{\textcolor{ForestGreen}{+.04}}$}
& {\bf $4.09_{\textcolor{ForestGreen}{+.06}}$}
& {\bf $4.09_{\textcolor{ForestGreen}{+.06}}$}  \\

HPSv2~\cite{hpsv2}  & {\bf $4.57_{\textcolor{ForestGreen}{+.06}}$} 
& {\bf $4.60_{\textcolor{ForestGreen}{+.09}}$}
& {\bf $3.80_{\textcolor{ForestGreen}{+.03}}$}
& {\bf $3.78_{\textcolor{ForestGreen}{+.01}}$}
& {\bf $4.07_{\textcolor{ForestGreen}{+.04}}$}
& {\bf $4.07_{\textcolor{ForestGreen}{+.04}}$}  \\
VQ2~\cite{vq2} & {\bf $4.54_{\textcolor{ForestGreen}{+.03}}$} 
& {\bf $4.55_{\textcolor{ForestGreen}{+.04}}$}
& {\bf $3.79_{\textcolor{ForestGreen}{+.02}}$}
& {\bf $3.79_{\textcolor{ForestGreen}{+.02}}$} 
& {\bf $4.05_{\textcolor{ForestGreen}{+.02}}$} 
& {\bf $4.06_{\textcolor{ForestGreen}{+.03}}$}  \\
Davidsonian~\cite{davidsonian} & {\bf $4.56_{\textcolor{ForestGreen}{+.05}}$} 
& {\bf $4.61_{\textcolor{ForestGreen}{+.10}}$}
& {\bf $3.83_{\textcolor{ForestGreen}{+.05}}$}
& {\bf $3.84_{\textcolor{ForestGreen}{+.07}}$} 
& {\bf $4.09_{\textcolor{ForestGreen}{+.06}}$} 
& {\bf $4.12_{\textcolor{ForestGreen}{+.09}}$}  \\
\midrule
{\bf VQAScore} & {\bf 4.59$_{\textcolor{ForestGreen}{+.08}}$} & {\bf 4.62$_{\textcolor{ForestGreen}{+.11}}$} & {\bf 3.92$_{\textcolor{ForestGreen}{+.15}}$} & {\bf 4.05$_{\textcolor{ForestGreen}{+.28}}$} & {\bf 4.16$_{\textcolor{ForestGreen}{+.13}}$} & {\bf 4.25$_{\textcolor{ForestGreen}{+.22}}$}  \\
\bottomrule[1.2pt]
\end{NiceTabular}
} & 
\scalebox{1.0}{
\begin{NiceTabular}{lcccccc}
    \CodeBefore
      \rectanglecolor{softgray}{3-1}{4-7}
      \rectanglecolor{softgreen}{11-1}{11-7}
    \Body
\toprule[1.2pt]
\multirow{2}{*}{\textbf{Method}} & \multicolumn{2}{c}{{\bf Basic}}  & \multicolumn{2}{c}{{\bf Advanced}} & \multicolumn{2}{c}{{\bf Overall}}\\ \cmidrule(r){2-3} \cmidrule{4-5}\cmidrule(l){6-7}
& {\textbf{\small 3 Imgs}} & \textbf{\small 9 Imgs} & \textbf{\small 3 Imgs} & {\textbf{\small 9 Imgs}} & \textbf{\small 3 Imgs} & \textbf{\small 9 Imgs} \\
\midrule
Random & $3.80 $ & $3.80$ & $3.02$ & $3.02 $ & $3.30$ & $3.30$ \\
Human (Oracle) & {\bf $4.17_{\textcolor{ForestGreen}{+.37}}$} 
& {\bf $4.41_{\textcolor{ForestGreen}{+.61}}$}
& {\bf $3.38_{\textcolor{ForestGreen}{+.36}}$}
& {\bf $3.70_{\textcolor{ForestGreen}{+.68}}$}
& {\bf $3.66_{\textcolor{ForestGreen}{+.36}}$}
& {\bf $3.95_{\textcolor{ForestGreen}{+.65}}$}  \\

\midrule
CLIPScore~\cite{clipscore}& {\bf $3.86_{\textcolor{ForestGreen}{+.06}}$}
& {\bf $3.92_{\textcolor{ForestGreen}{+.12}}$}
& {\bf $3.06_{\textcolor{ForestGreen}{+.04}}$}
& {\bf $3.06_{\textcolor{ForestGreen}{+.04}}$}
& {\bf $3.34_{\textcolor{ForestGreen}{+.04}}$} 
& {\bf $3.37_{\textcolor{ForestGreen}{+.07}}$}  \\

ImageReward~\cite{imagereward} & {\bf 3.94$_{\textcolor{ForestGreen}{+.14}}$} 
& {\bf $3.96_{\textcolor{ForestGreen}{+.16}}$}
& {\bf $3.10_{\textcolor{ForestGreen}{+.08}}$}
& {\bf $3.15_{\textcolor{ForestGreen}{+.13}}$} 
& {\bf $3.40_{\textcolor{ForestGreen}{+.10}}$} 
& {\bf $3.44_{\textcolor{ForestGreen}{+.14}}$}  \\

PickScore~\cite{pickscore} & {\bf $3.94_{\textcolor{ForestGreen}{+.14}}$} & {\bf $4.02_{\textcolor{ForestGreen}{+.22}}$} & {\bf $3.13_{\textcolor{ForestGreen}{+.11}}$} & {\bf $3.17_{\textcolor{ForestGreen}{+.15}}$}  & {\bf $3.42_{\textcolor{ForestGreen}{+.12}}$}  & {\bf $3.47_{\textcolor{ForestGreen}{+.17}}$}  \\

HPSv2~\cite{hpsv2}  & {\bf $3.91_{\textcolor{ForestGreen}{+.11}}$} & {\bf $3.99_{\textcolor{ForestGreen}{+.19}}$} & {\bf $3.11_{\textcolor{ForestGreen}{+.09}}$}  & {\bf $3.15_{\textcolor{ForestGreen}{+.13}}$} & {\bf $3.39_{\textcolor{ForestGreen}{+.09}}$} & {\bf $3.45_{\textcolor{ForestGreen}{+.15}}$}  \\

VQ2~\cite{vq2} & {\bf $3.83_{\textcolor{ForestGreen}{+.03}}$} 
& {\bf $3.88_{\textcolor{ForestGreen}{+.08}}$}
& {\bf $3.06_{\textcolor{ForestGreen}{+.04}}$}
& {\bf $3.08_{\textcolor{ForestGreen}{+.06}}$} 
& {\bf $3.33_{\textcolor{ForestGreen}{+.03}}$} 
& {\bf $3.37_{\textcolor{ForestGreen}{+.07}}$}  \\

Davidsonian~\cite{davidsonian} & {\bf $3.85_{\textcolor{ForestGreen}{+.05}}$} 
& {\bf $3.88_{\textcolor{ForestGreen}{+.08}}$}
& {\bf $3.06_{\textcolor{ForestGreen}{+.04}}$}
& {\bf $3.11_{\textcolor{ForestGreen}{+.09}}$} 
& {\bf $3.34_{\textcolor{ForestGreen}{+.04}}$} 
& {\bf $3.39_{\textcolor{ForestGreen}{+.09}}$}  \\
\midrule
{\bf VQAScore} & {\bf 3.94$_{\textcolor{ForestGreen}{+.14}}$} & {\bf 4.06$_{\textcolor{ForestGreen}{+.26}}$} & {\bf 3.15$_{\textcolor{ForestGreen}{+.13}}$} & {\bf 3.29$_{\textcolor{ForestGreen}{+.27}}$} & {\bf 3.43$_{\textcolor{ForestGreen}{+.13}}$} & {\bf 3.56$_{\textcolor{ForestGreen}{+.26}}$}  \\
\bottomrule[1.2pt]
\end{NiceTabular}
} \\
\text{\large (a) Improving DALL-E 3 by image ranking} & \text{\large (b) Improving SD-XL by image ranking}
\end{tabular}
}
\label{tab:ranking_all}
\end{table*}

\clearpage
\section{Goodhart's Law Still Applies}
\label{sec:limitation}

\begin{quote}
\raggedright %
{\em When a measure becomes a target, it ceases to be a good measure.}\\
\hfill--- Marilyn Strathern~\cite{strathern1997improving} %
\end{quote}

\noindent This quote conveys the essence of Goodhart's Law~\cite{goodhart1984problems, goodhart2015goodhart}: an over-optimized metric inevitably loses its effectiveness. This phenomenon is well-documented in fields such as machine learning~\cite{teney2020value, hsia2023goodhart}, economics~\cite{danielsson2002emperor, goodhart1984problems}, and education~\cite{biagioli2016watch, koltun2021h}. Acknowledging that VQAScore is also subject to this law, we examine its limitations as an automated metric and suggest avenues for future improvements.

\begin{figure}[h!]
\centering
    \scalebox{0.96}{
        \begin{tabular}{ccc}
           (a) Too many objects & (b) Fine-grained visual details & (c) Linguistic ambiguity \\
           \includegraphics[width=0.31\textwidth]{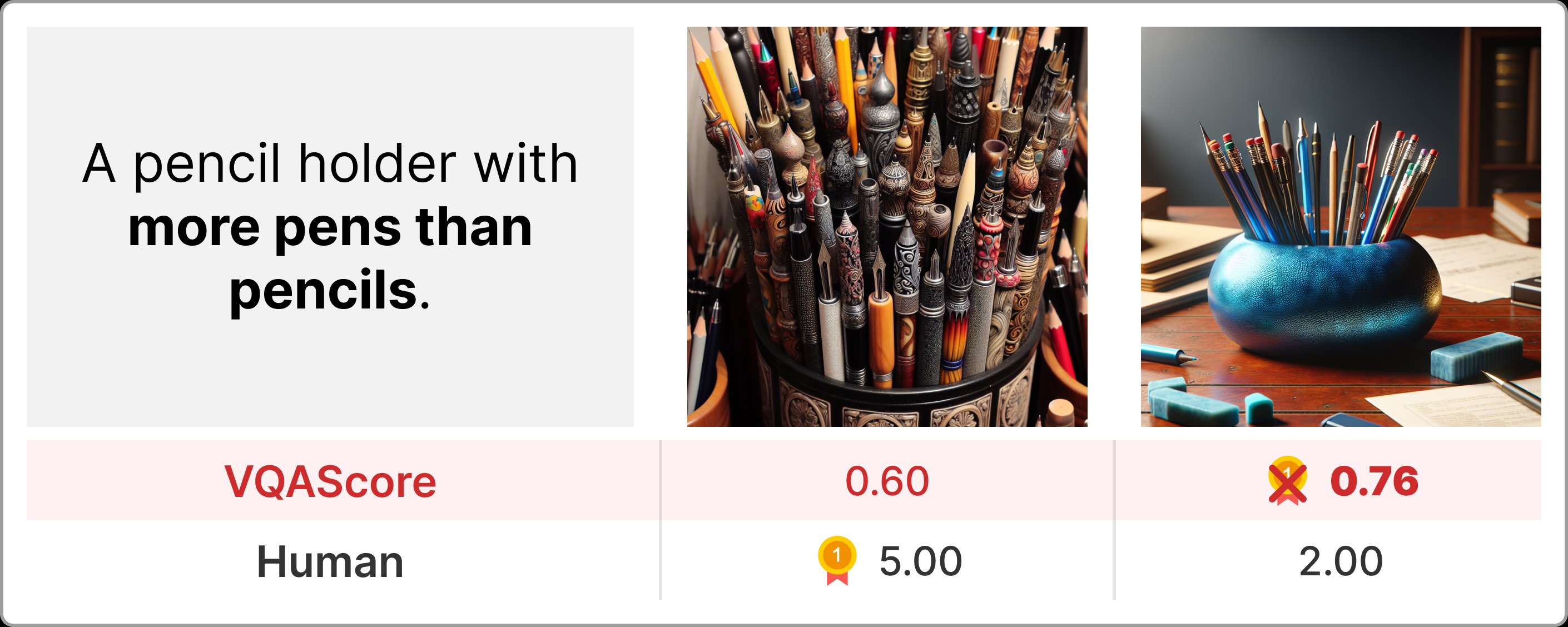} &  \includegraphics[width=0.31\textwidth]{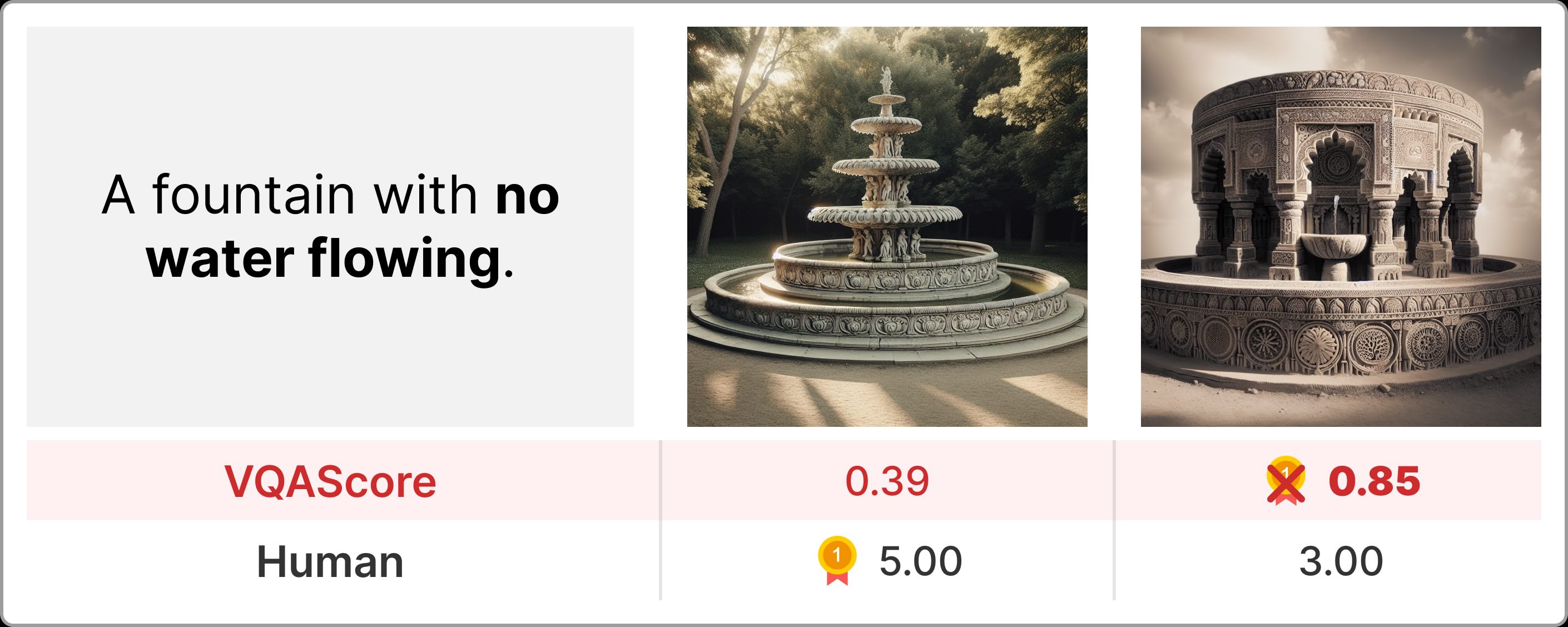} &\includegraphics[width=0.31\textwidth]{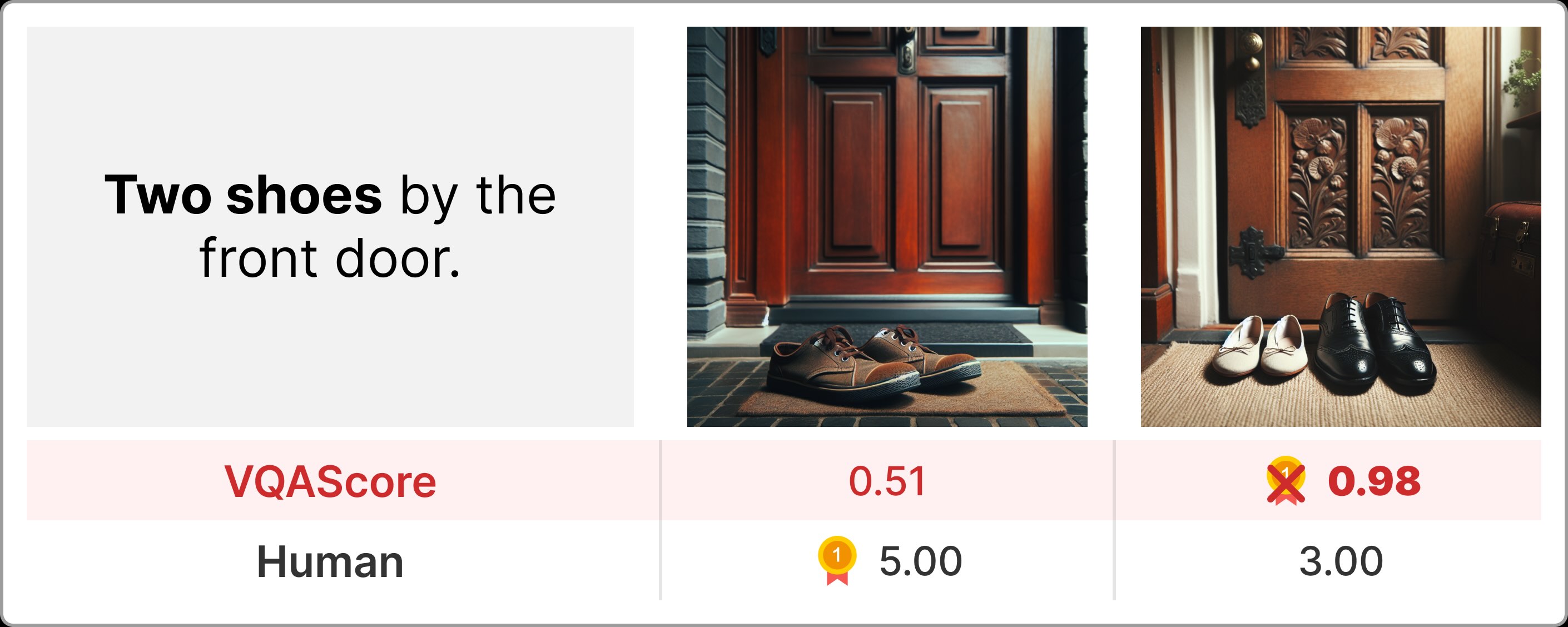} \\
           \includegraphics[width=0.31\textwidth]{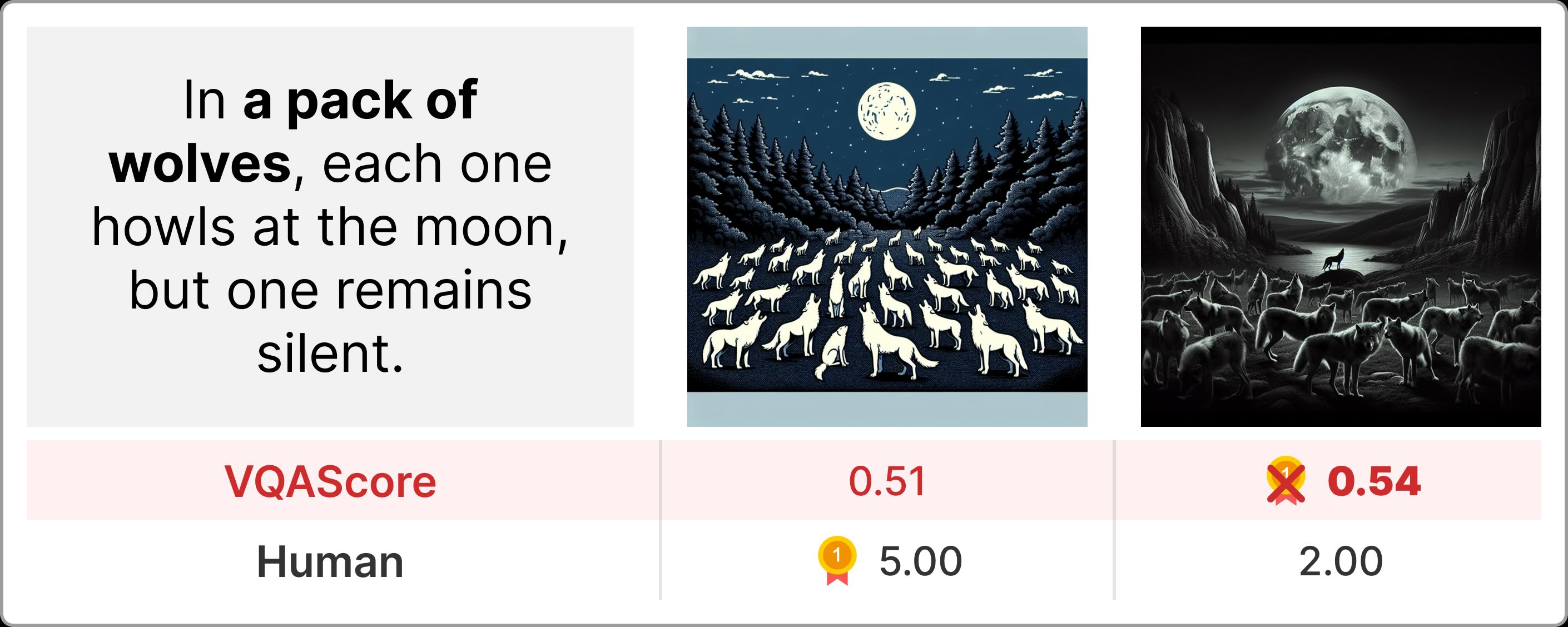} & \includegraphics[width=0.31\textwidth]{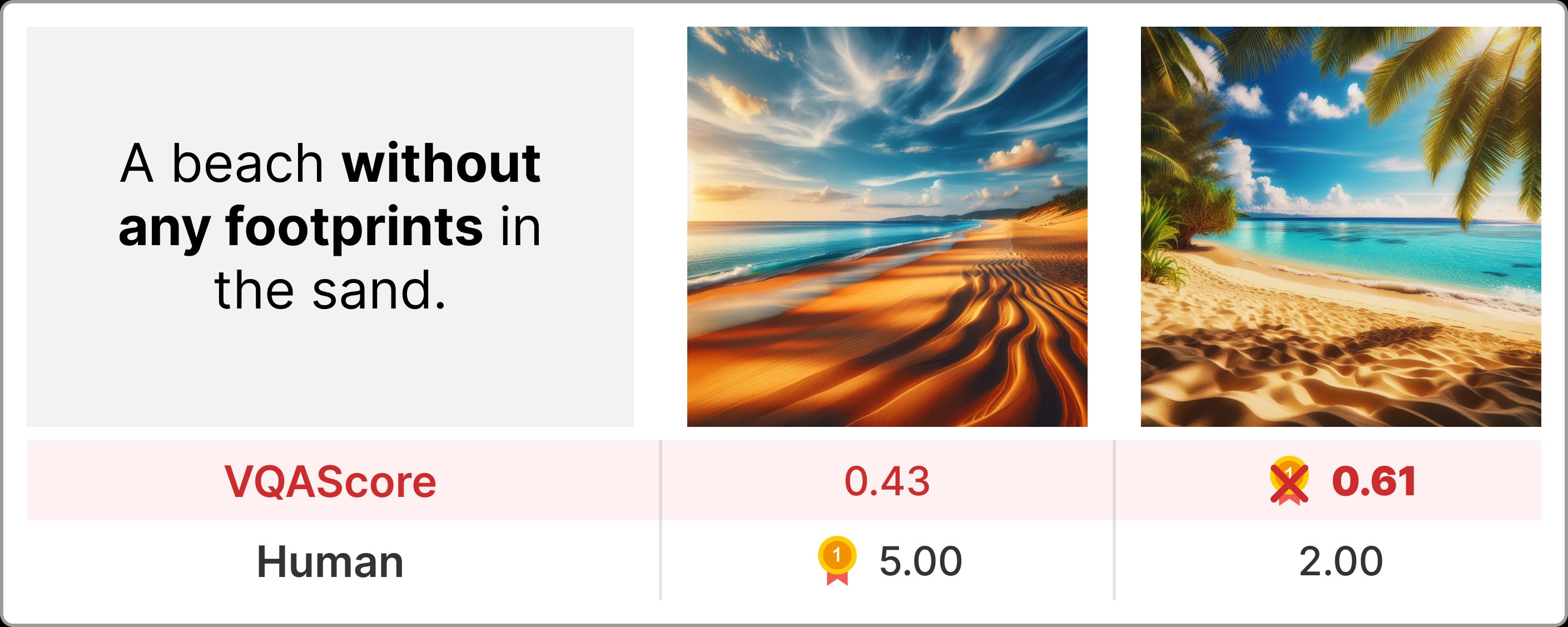} &\includegraphics[width=0.31\textwidth]{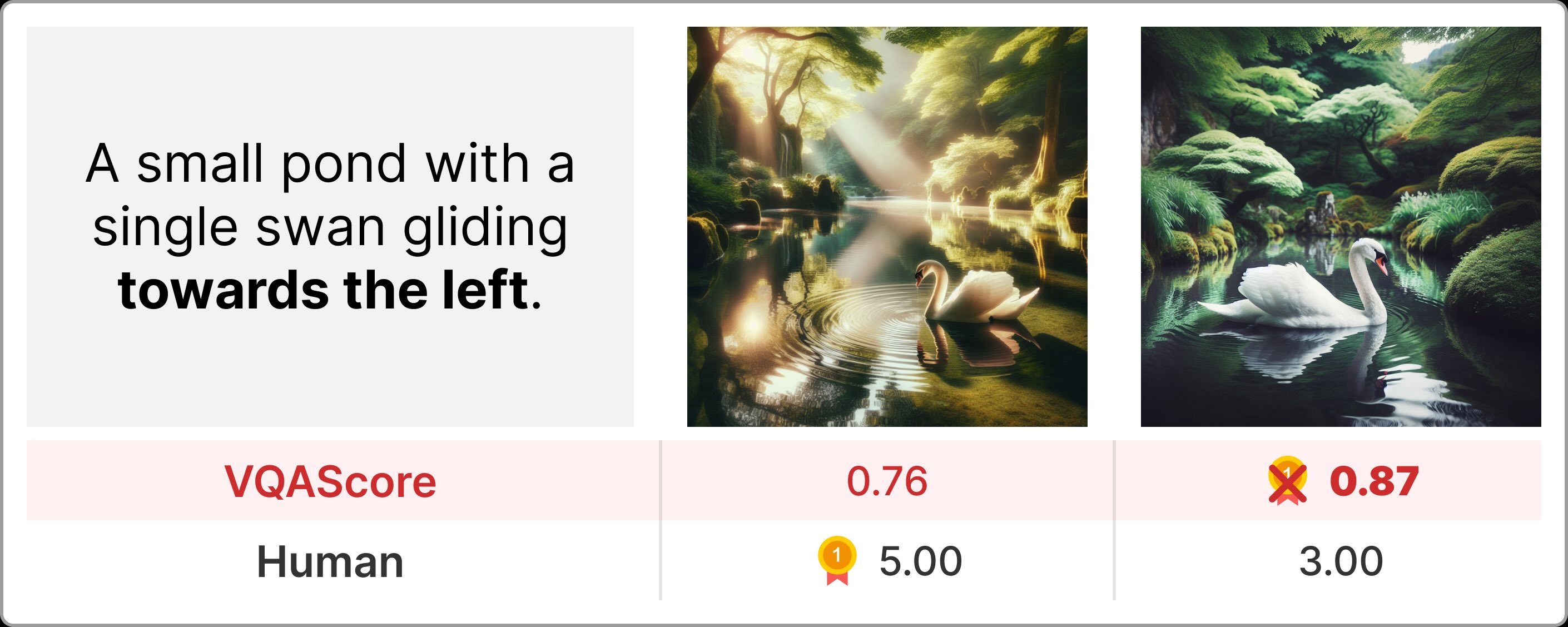} 
        \end{tabular}
    }
    \caption{\small {\bf Limitations of VQAScore (please zoom into the figures for a detailed view).} We identify three failure cases of VQAScore (based on CLIP-FlanT5). (a) While VQAScore can reasonably count objects in small quantities, it struggles with larger numbers. (b) VQAScore can overlook small visual details, such as entities that occupy only a small portion of the image. (c) VQAScore may not understand ambiguous prompts, misinterpreting ``two shoes'' as ``two pairs of shoes'',  or ``towards the left (of the viewer)'' as ``towards the left (of the swan)''.}
    \label{fig:vqascore_limitations}
\end{figure}

{\bf Limitations of VQAScore.} We conduct a qualitative study by manually examining samples  
 where VQAScore and human ratings disagree. \autoref{fig:vqascore_limitations} identifies three failure cases: (1) miscounting when there are too many objects, (2) overlooking fine-grained visual details, and (3) misinterpreting linguistic ambiguity. We posit that VQA models with higher image resolution~\cite{shi2024we} and more capable language models~\cite{ul2, gpt4} may improve on these challenging aspects. Despite these mild limitations, we strongly recommend adopting VQAScore as a more reliable alternative to CLIPScore, which has already ceased to be an effective metric~\cite{kamath2023text, aro, lin2024revisiting}. We believe VQAScore also serves well as a reproducible supplement to non-reproducible human studies~\cite{otani2023toward}.

\section{Conclusion}
\label{sec:conclusion}
{\bf Limitations and future work.} Currently, GenAI-Bench does not evaluate several vital aspects of generative models~\cite{lee2023holistic, parashar2024neglected, evalcrafter, wu2024gpt}, such as toxicity, bias, aesthetics, and video motion. Future work may also incorporate other interesting aspects of visual generation, such as mixed media, optical effects, reflection, and world knowledge, as explored in datasets such as DOCCI~\cite{docci}. Although our ranking-based approach is effective, future work may explore white-box finetuning techniques~\cite{black2023training, prabhudesai2023aligning, wu2024deep} for more efficient inference.

{\bf Summary.} We have conducted an extensive human study with GenAI-Bench, focusing on both compositional text-to-visual generation and automated evaluation metrics. We show a straightforward ranking-based method that improves the prompt alignment of black-box generative models. By discussing Goodhart's Law, we hope to encourage further research into automated evaluation techniques, which is essential to the scientific progression of this field.

\clearpage
{
    \small
    \bibliographystyle{ieeenat_fullname}
    \bibliography{main}
}

\newpage
\clearpage

{
    \centering
    \Large
    \textbf{GenAI-Bench: Evaluating and Improving Compositional Text-to-Visual Generation} \\ \vspace{0.5em} {Supplementary Material}\\
    \large
}

\appendix

\vspace{-0.1in}

\section*{}
\begin{center}
    \emph{\bf \em \large Outline}
\end{center}
{This document supplements the main paper with benchmark and method details. Below is the outline:
\begin{itemize}
\item {\bf Section \ref{sec:additional_examples}} presents additional qualitative studies.
\item {\bf Section \ref{sec:skill_details}} details GenAI-Bench's skill taxonomy.
\item {\bf Section \ref{sec:genai_bench_details}} describes how we collect GenAI-Bench.
\item {\bf Section \ref{sec:method_details}} discusses other baseline methods.

\end{itemize}
}

\section{Additional Examples}
\label{sec:additional_examples}
\begin{figure*}[htbp!]
\centering
    \scalebox{0.9}{
        \begin{tabular}{c@{\hspace{1pt}}c}
           \includegraphics[width=0.5\textwidth]{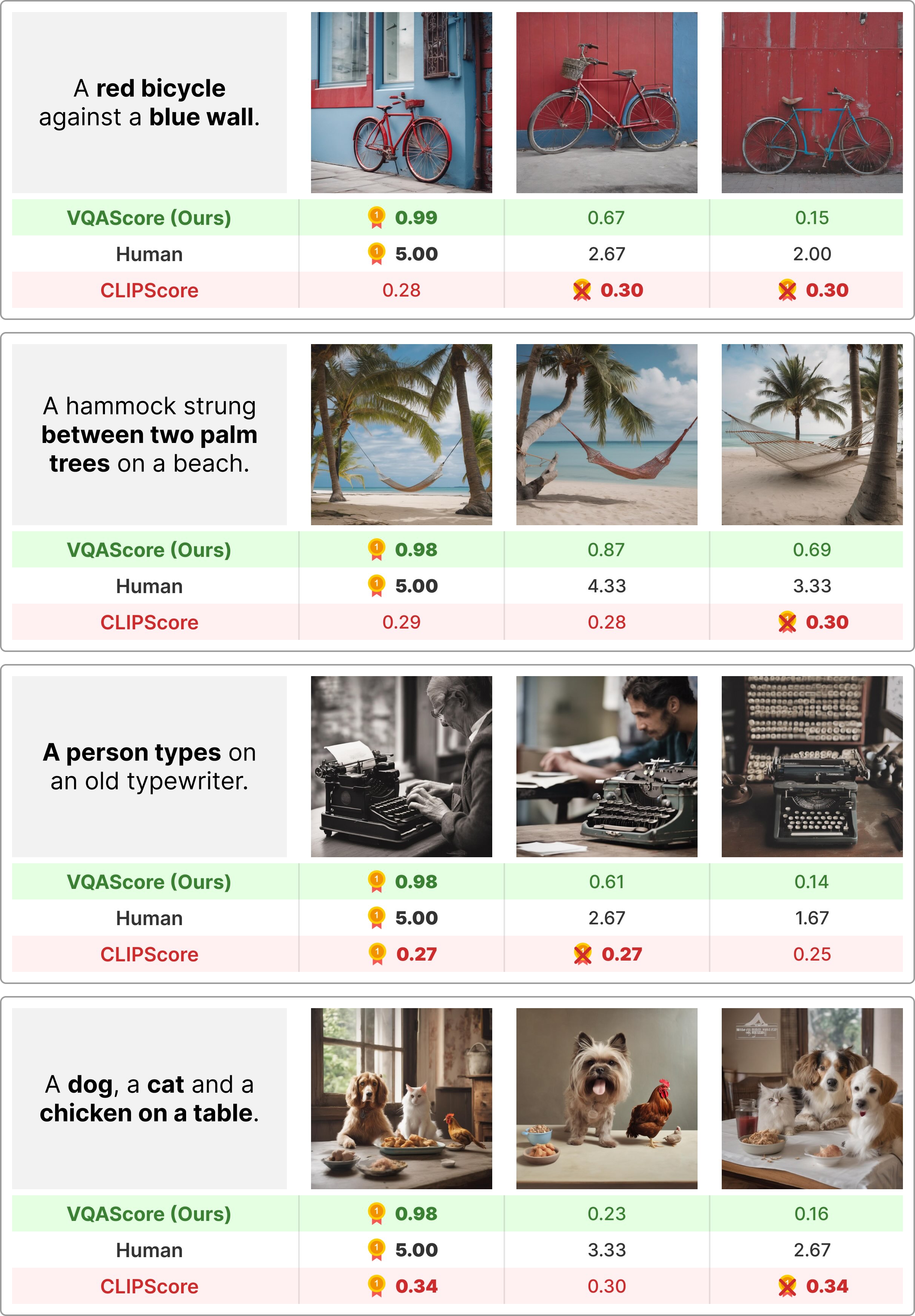} & \includegraphics[width=0.5\textwidth]{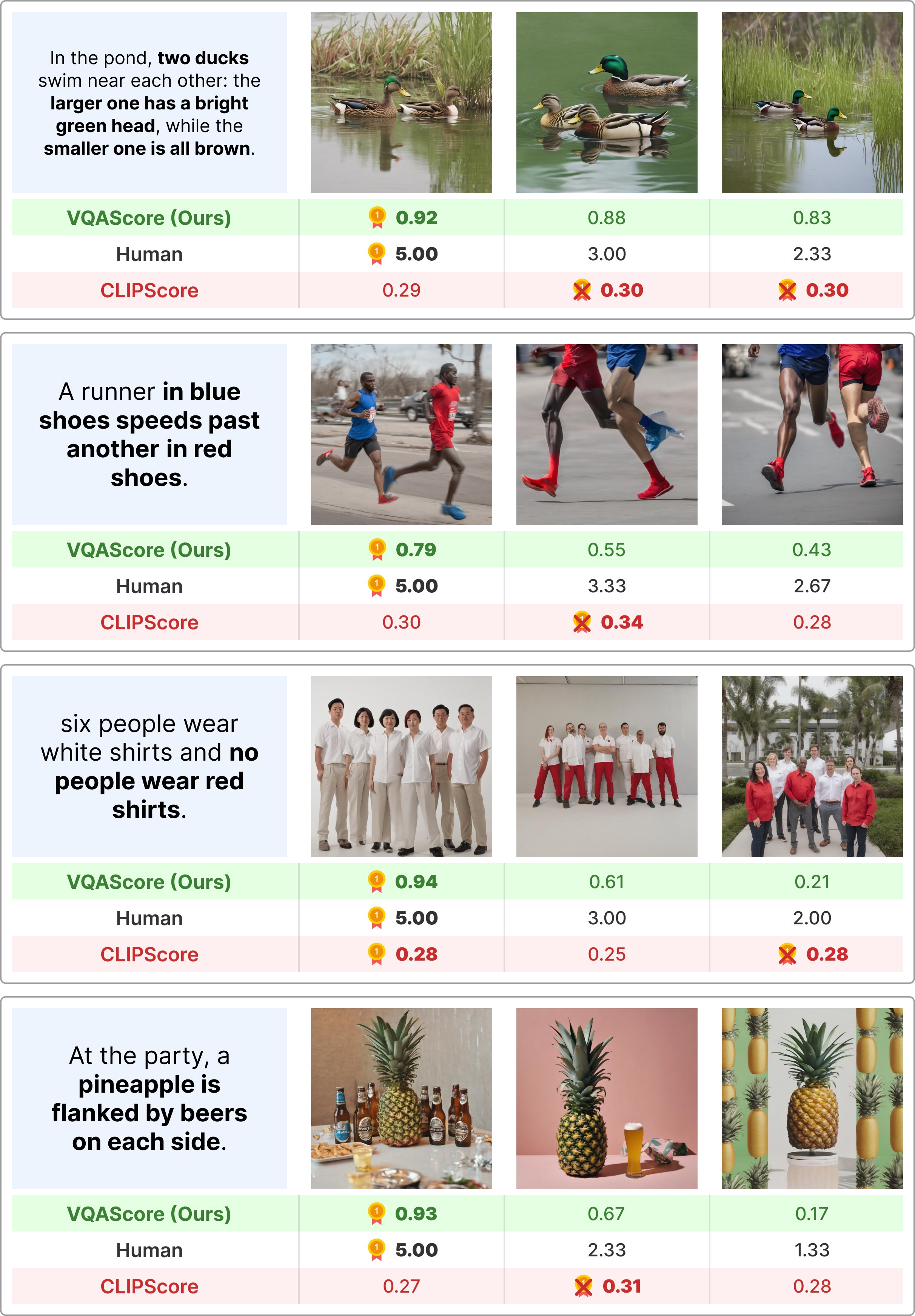} \\
        \end{tabular}
    }
    \caption{\small {\bf Ranking SD-XL generated images with VQAScore and CLIPScore.} VQAScore outperforms CLIPScore in ranking candidate images generated by SD-XL, particularly for advanced prompts that involve complex visio-linguistic reasoning. }
    \label{fig:ranking_sdxl}
\end{figure*}

{\bf Ranking SD-XL images by VQAScore.} \autoref{fig:ranking_sdxl} shows that ranking by VQAScore can also improve the prompt alignment of SD-XL using only its image generation API. We encourage future work to explore other white-box techniques for finetuning~\cite{sun2023dreamsync, ruiz2023dreambooth, wallace2023diffusion, lin2023crossmodal, lee2023aligning, guo2024versat2i, li2024reward}.

\clearpage
\section{Skill taxonomy of GenAI-Bench}
\label{sec:skill_details}
We now detail the evaluated skills of GenAI-Bench.

{\bf Skill definitions.} Previous benchmarks for text-to-visual generation~\cite{parti, imagen, tifa, dalleval, huang2023t2i} primarily focus on generating {\em basic} objects, attributes, relations, and scenes. While these ``basic'' visual compositions still pose challenges, real-world user prompts often introduce greater complexity, such as higher-order reasoning beyond basic compositions. For example, while existing benchmarks focus only on counting objects~\cite{tifa, parti}, real-world prompts often require counting complex object-attribute-relation compositions, like ``{\tt one person wearing a white shirt and the other five wearing blue shirts}''. To this end, after reviewing relevant literature~\cite{winoground, midjourney, huang2023t2i, parti} and discussing with professional designers, we introduce a set of compositional reasoning skills and categorize them into ``basic'' and ``advanced'', where the latter can build upon the former. For logical reasoning, we focus on ``negation'' and ``universality''. We find these are the two most common types of logic in user prompts. Other logical operators such as conjunction~\cite{liu2022compositional} are not included because the logical ``AND'' is usually implied in prompts that involve multiple objects, and ``OR'' is rarely useful as designers tend to specify a more precise prompt. Lastly, we plan to evaluate image styles in future work. We detail the definitions for ``basic'' skills in \autoref{tab:basic_skills} and ``advanced'' skills in \autoref{tab:advanced_skills}. 

\begin{table*}[h!]
\centering
\caption{\textbf{Skill definitions and examples for basic compositions.} }
\scalebox{0.8}{
\begin{NiceTabular}{l M{0.4\linewidth} M{0.5\linewidth}}
\CodeBefore
      \rectanglecolor{softgray}{2-1}{15-3}
    \Body
\toprule[1.2pt]
 \textbf{Skill Type} & \textbf{Definition} & \textbf{Examples} \\ \midrule
 \multicolumn{3}{c}{\bf Basic Compositions}\\
 {\fontfamily{cmtt}\selectfont Object} & {Basic entities within an image, such as person, animal, food, items, vehicles, or text symbols (e.g., ``A'', ``1+1'').} & {\it a {\bf dog}, a {\bf cat} and a {\bf chicken} on a {\bf table}; a young {\bf man} with a green {\bf bat} and a blue {\bf ball}; a '{\bf No Parking}' sign on a busy street.} \\\\
 {\fontfamily{cmtt}\selectfont Attribute} & {Visual attributes (properties) of entities, such as color, material, emotion, size, shape, age, gender, state, and so on.} & {\it a {\bf silver} spoon lies to the left of a {\bf golden} fork on a {\bf wooden} table; a {\bf green} pumpkin is smiling {\bf happily}, a {\bf red} pumpkin is sitting {\bf sadly}.} \\\\
 {\fontfamily{cmtt}\selectfont Scene} & {Backgrounds or settings of an image, such as weather and location.} & {\it A child making a sandcastle on a {\bf beach in a cloudy day}; a grand fountain surrounded by historic buildings in a {\bf town square}.} \\\\
 {\fontfamily{cmtt}\selectfont Spatial Relation} & {Physical arrangements of multiple entities relative to each other, e.g., on the right, on top, facing, towards, inside, outside, near, far, and so on.} & {\it a bustling city street, a neon 'Open 24 Hours' sign glowing {\bf above} a small diner; a teacher standing {\bf in front of} a world map in a classroom; tea steams {\bf in} a cup, {\bf next to} a closed diary with a pen resting {\bf on} its cover.} \\\\
 {\fontfamily{cmtt}\selectfont Action Relation} & {Action interactions between entities, e.g., pushing, kissing, hugging, hitting, helping, and so on.} & {\it a group of children {\bf playing} on the beach; a boat {\bf glides} across the ocean, dolphins {\bf leaping} beside it and seagulls {\bf soaring} overhead.} \\\\
 {\fontfamily{cmtt}\selectfont Part Relation} & {Part-whole relationships between entities -- one entity is a component of another, such as body part, clothing, and accessories.} & {\it a pilot {\bf with aviator sunglasses}; a baker {\bf with a cherry pin on a polka dot apron}.; a young lady {\bf wearing a T-shirt} puts {\bf her hand} on a {\bf puppy's head}.}  \\
\bottomrule[1.2pt]
\end{NiceTabular}
}
\label{tab:basic_skills}
\end{table*}

\begin{table*}[h!]
\centering
\caption{\textbf{Skill definitions and examples for advanced compositions.} }
\scalebox{0.8}{
\begin{NiceTabular}{l M{0.4\linewidth} M{0.5\linewidth}}
\CodeBefore
      \rectanglecolor{softblue}{2-1}{15-3}
    \Body
\toprule[1.2pt]
 \textbf{Skill Type} & \textbf{Definition} & \textbf{Examples} \\ \midrule
 \multicolumn{3}{c}{\bf Advanced Compositions}\\
 {\fontfamily{cmtt}\selectfont Counting} & {Determining the quantity, size, or volume of entities, e.g., objects, attribute-object pairs, and object-relation-object triplets.} & {\it {\bf two} cats playing with a {\bf single} ball; {\bf five} enthusiastic athletes and {\bf one} tired coach; {\bf one} pirate ship sailing through space, crewed by {\bf five} robots; {\bf three} pink peonies and {\bf four} white daisies in a garden.} \\\\
 {\fontfamily{cmtt}\selectfont Differentiation} & {Differentiating objects within a category by their attributes or relations, such as distinguishing between ``old'' and ``young'' people by age, or ``the cat on top of the table'' versus ``the cat under the table'' by their spatial relations.} & {\it {\bf one cat} is sleeping on the table and {\bf the other} is playing under the table; there are two men in the living room, {\bf the taller one} to the left of {\bf the shorter one}; a notebook lies open in the grass, with sketches on {\bf the left page} and blank space {\bf on the right}; there are two shoes on the grass, {\bf the one without laces} looks newer than {\bf the one with laces}.} \\\\
 {\fontfamily{cmtt}\selectfont Comparison} & {Comparing characteristics like number, attributes, area, or volume between entities.} & {\it there are {\bf more} people standing than sitting; between the two cups on the desk, the {\bf taller} one holds {\bf more} coffee than the {\bf shorter} one, which is half-empty; three little boys are sitting on the grass, and the boy in the middle looks the {\bf strongest}.} \\\\
 {\fontfamily{cmtt}\selectfont Negation} & {Specifying the absence or contradiction of elements, as indicated by ``no'', ``not'', or ``without'', e.g., entities not present or actions not taken.} & {\it  a bookshelf with {\bf no} books, only picture frames.; a person with short hair is crying while a person with long hair {\bf is not}; a smiling girl with short hair and {\bf no} glasses}; a cute dog {\bf without} a collar. \\\\
 {\fontfamily{cmtt}\selectfont Universality} & {Specifying when every member of a group shares a specific attribute or is involved in a common relation, indicated by words like ``every'', ``all'', ``each'', ``both''.} & {\it in a room, {\bf all} the chairs are occupied except one; a bustling kitchen where {\bf every} chef is preparing a dish; in a square, several children are playing, {\bf each} wearing a red T-shirt; a table laden with apples and bananas, where {\bf all} the fruits are green; the little girl in the garden has roses in {\bf both} hands.} \\
\bottomrule[1.2pt]
\end{NiceTabular}
}
\label{tab:advanced_skills}
\end{table*}

{\bf Skill coverage in benchmarks.} We find the skill categorization in benchmarks like PartiPrompt~\cite{parti} to be somewhat confusing. For instance, PartiPrompt introduces two categories ``{\em complex}'' and ``{\em fine-grained detail}''. The former refers to ``{\em ...fine-grained, interacting details or relationships between multiple participants}'', while the latter refers to ``{\em ...attributes or actions of entities or objects in a scene}''. Upon closer examination, the categorization of spatial, action, and part relations into these categories are arbitrary. To address this, we attempt to compare the skill coverage across all benchmarks by our unified set of compositional reasoning skills. For benchmarks (PartiPrompt/T2I-CompBench) with pre-defined skill categories, we map their skills to our definitions. For the other benchmarks that do not have a comprehensive skill set, we manually annotate a random subset of samples. Finally, we calculate the skill proportions in each benchmark, identifying skills that constitute more than 2\% as genuinely present.

\section{GenAI-Bench}
\label{sec:genai_bench_details}
This section describes how we collect GenAI-Bench.

{\bf Details of GenAI-Bench.} GenAI-Bench consists of 1,600 diverse prompts. To ensure our prompts reflect real-world applications, we collaborate with graphic designers who regularly use text-to-visual tools like Midjourney~\cite{midjourney}. First, we collaborate with them to refine our skill taxonomy, identifying practical skills that current models still struggle with. They then collect compositional prompts relevant to their professional needs. To avoid copyright issues, we advise them to write prompts about generic subjects. We provide designers with sample prompts from existing benchmarks like PartiPrompt for inspiration and encourage the use of ChatGPT to brainstorm prompt variants across diverse visual domains. Crucially, these designers ensure that the prompts are {\em objective}. This contrasts with \cite{huang2023t2i}, whose prompts are almost entirely auto-generated. For example, in the ``{\em texture}'' category of T2I-CompBench, an overwhelming 40\% of the 1000 auto-generated prompts use ``metallic'' as the attribute, which limits their diversity. Other T2I-CompBench's prompts generated by ChatGPT often contain subjective phrases. For instance, in the prompt ``{\tt the delicate, fluttering wings of the butterfly signaled the arrival of spring, a natural symbol of rebirth and renewal}'', the ``rebirth and renewal'' can convey different meanings to different people. Similarly, in ``{\tt the soft, velvety texture of the rose petals felt luxurious against the fingertips, a romantic symbol of love and affection}'', the ``love and affection'' is open to diverse interpretations. Thus, we guide the designers to avoid such prompts. Lastly, each prompt in GenAI-Bench is tagged with all its evaluated skills. In total, we collect over 5,000 human-verified tags with a balanced distribution of ``basic'' and ``advanced'' skills.

{\bf Collecting human ratings.} We evaluate six text-to-image models: Stable Diffusion~\cite{stablediffusion} (SD v2.1, SD-XL, SD-XL Turbo), DeepFloyd-IF~\cite{deepfloyd}, Midjourney v6~\cite{midjourney}, DALL-E 3~\cite{dalle3}; along with four text-to-video models: ModelScope~\cite{modelscope}, Floor33~\cite{floor33}, Pika v1~\cite{pikalab}, Gen2~\cite{gen2}. Due to the lack of APIs for Floor33~\cite{floor33}, Pika v1~\cite{pikalab}, and Gen2~\cite{gen2}, we manually download videos from their websites. For image generative models, we generate images using all 1,600 GenAI-Bench prompts. We use a coreset of 800 prompts to collect videos for the four video models. The same 800 prompts are used to collect the GenAI-Rank benchmark in the main paper. In total, we collect over 80,000 human ratings, greatly exceeding the scale of human annotations in previous work~\cite{tifa, davidsonian}, e.g., TIFA160 collected 2,400 ratings. We pay the local minimum wage of 12 dollars per hour for a total of about 800 annotator hours.

{\bf GenAI-Bench performance.} We detail the performance of the ten image and video generative models across all skills in \autoref{tab:performance_breakdown}. Both humans and VQAScores rate DALL-E 3~\cite{dalle3} higher than the other models in nearly all skills, except for negation. In addition, prompts requiring ``advanced'' compositions are rated significantly lower by both humans and VQAScores, with negation being the most challenging skill. Lastly, current video models do not perform as well as image models, suggesting room for improvement.

{\bf Performance of automated metrics across skill dimensions.} To provide a more comprehensive comparison between VQAScore and other metrics, we report Pairwise Accuracy across basic and advanced skill dimensions on GenAI-Bench (Image) in \autoref{tab:comprehenseive_comparison_with_metrics}.

\begin{table*}[h!]
\centering
\renewcommand{\arraystretch}{1.3}
\caption{{\bf Performance breakdown on GenAI-Bench.} We present the averaged human ratings and VQAScores (based on CLIP-FlanT5) for ``basic'' and ``advanced'' prompts. Human ratings use a 1-5 Likert scale, and VQAScore ranges from 0 to 1, with higher scores indicating better performance for both. Generally, both human ratings and VQAScores favor DALL-E 3 over other models, with DALL-E 3 preferred across almost all skills except for negation. We find that ``advanced'' prompts that require higher-order reasoning present significant challenges. For instance, the state-of-the-art DALL-E 3 receives a remarkable average human rating of 4.3 for ``basic'' prompts, indicating the images and prompts range from ``{\em having a few minor discrepancies}'' to ``{\em matching exactly}''. However, it scores only 3.4 for ``advanced'' prompts, suggesting ``{\em several minor discrepancies}''. In addition, video models receive significantly lower scores than image models. Overall, VQAScores closely match human ratings.
}
\scalebox{0.6}{
\begin{tabular}{c@{\hspace{8pt}}c}

  \begin{NiceTabular}{lccccc|c}
        \CodeBefore
        \Body
    \toprule[1.2pt]
    \multirow{2}{*}{\textbf{Method}} & \multirow{2}{*}{\bf Attribute} & \multirow{2}{*}{\bf Scene} & \multicolumn{3}{c}{\bf Relation} & \multirow{2}{*}{\bf Avg}   \\
    \cmidrule{4-6}
    &  &  &  Spatial & Action & Part   \\
    \midrule
    \multicolumn{5}{l}{\textit{Image models}}\\
    SD v2.1 & 3.2 & 3.4 & 3.1 & 3.2 & 2.9 & 3.2\\
    SD-XL Turbo & 3.5 & 3.6 & 3.4& 3.4 & 3.2 & 3.5 \\
    SD-XL & 3.6 & 3.8 & 3.5 & 3.6 & 3.4 & 3.6\\
    DeepFloyd-IF & 3.6 & 3.7 & 3.6 & 3.6 & 3.4 & 3.6 \\
    Midjourney v6 & 4.0 & 4.1 & 4.0 & 4.1 & 3.8 & 4.0\\
    DALL-E 3 & 4.3 & 4.4 & 4.2 & 4.3 & 4.2 & 4.3\\
    \midrule
    \multicolumn{5}{l}{\textit{Video models}}\\
    ModelScope & 3.1 & 3.1 & 2.8 & 3.0 & 3.1 & 3.0\\
    Floor33 & 3.2 & 3.2 & 2.9 & 3.2 & 3.1 & 3.1\\
    Pika v1 & 3.4 & 3.4 & 3.1 & 3.3 & 3.2 & 3.3\\
    Gen2 & 3.6 & 3.7 & 3.4 & 3.6 & 3.6 & 3.6\\
    \bottomrule[1.2pt]
    \end{NiceTabular}
    
         & 
           \begin{NiceTabular}{lccccc|c}
        \CodeBefore
        \Body
    \toprule[1.2pt]
    \multirow{2}{*}{\textbf{Method}} & \multirow{2}{*}{\bf Attribute} & \multirow{2}{*}{\bf Scene} & \multicolumn{3}{c}{\bf Relation} & \multirow{2}{*}{\bf Avg}   \\
    \cmidrule{4-6}
    &  &  &  Spatial & Action & Part   \\
    \midrule
    \multicolumn{5}{l}{\textit{Image models}}\\
    SD v2.1 & 0.75 & 0.79 & 0.73 & 0.73 & 0.71 & 0.75\\
    SD-XL Turbo & 0.81 & 0.82 & 0.78 & 0.79 & 0.78 & 0.80\\
    SD-XL & 0.82 & 0.85 & 0.80 & 0.80 & 0.81 & 0.82\\ 
    DeepFloyd-IF & 0.82 & 0.83 & 0.80 & 0.81 & 0.81 & 0.82\\
    Midjourney v6 & 0.86 & 0.88 & 0.86 & 0.87 & 0.85 & 0.86 \\
    DALL-E 3 & 0.91 & 0.91 & 0.90 & 0.90 & 0.91 & 0.90 \\
    \midrule
    \multicolumn{5}{l}{\textit{Video models}}\\
    ModelScope & 0.69 & 0.69 & 0.65 & 0.65 & 0.70 & 0.66\\
    Floor33 & 0.70 & 0.71 & 0.64& 0.66 & 0.67 & 0.67\\
    Pika v1 & 0.78 & 0.80 & 0.74 & 0.72 & 0.76 & 0.75\\
    Gen2 & 0.79 & 0.81 & 0.74 & 0.76 & 0.83 & 0.77\\
    \bottomrule[1.2pt]
    \end{NiceTabular}

    \\
  {\bf (a) Human ratings on ``basic'' prompts}   & {\bf (b) VQAScores on ``basic'' prompts} \\ \\

        \begin{NiceTabular}{lccccc|c}
        \CodeBefore
        \Body
    \toprule[1.1pt]
    \multirow{2}{*}{\textbf{Method}} &  \multirow{2}{*}{\bf Count} & \multirow{2}{*}{\bf Differ} & \multirow{2}{*}{\bf Compare} & \multicolumn{2}{c}{\bf Logical}  & \multirow{2}{*}{\bf Avg}   \\
    \cmidrule{5-6}
    &  &  & &  Negate & Universal   \\
    \midrule
    \multicolumn{5}{l}{\textit{Image models}}\\
    SD v2.1 & 2.8 & 2.5 & 2.6 & 2.9 & 3.2 & 2.9\\
    SD-XL Turbo & 2.9 & 2.6 & 2.6 & 2.9 & 3.3 & 3.0\\
    SD-XL & 3.0 & 2.7 & 2.6 & 2.9 & 3.3 & 3.0\\
    DeepFloyd-IF & 3.2 & 2.9 & 2.9 & 2.9& 3.5 & 3.1\\
    Midjourney v6 & 3.4 & 3.2 & 3.2 & 3.0 &3.7 & 3.4\\
    DALL-E 3 & 3.6 & 3.5 & 3.4 & 3.0 & 3.8 & 3.4\\
    \midrule
    \multicolumn{5}{l}{\textit{Video models}}\\
    ModelScope & 2.4 & 2.4 & 2.2 & 2.6 & 2.8 & 2.5\\
    Floor33 & 2.7 & 2.7 & 2.5 & 2.8 & 3.2 & 2.8\\
    Pika v1 & 2.7 & 2.7 & 2.6 & 2.9 & 3.3 & 2.9\\
    Gen2 & 2.8 & 2.7 & 2.6 & 2.9 & 3.3 & 2.9\\
    \bottomrule[1.1pt]
    \end{NiceTabular}

         & 
         
    \begin{NiceTabular}{lccccc|c}
        \CodeBefore
        \Body
    \toprule[1.1pt]
    \multirow{2}{*}{\textbf{Method}} &  \multirow{2}{*}{\bf Count} & \multirow{2}{*}{\bf Differ} & \multirow{2}{*}{\bf Compare} & \multicolumn{2}{c}{\bf Logical}  & \multirow{2}{*}{\bf Avg}   \\
    \cmidrule{5-6}
    &  &  & &  Negate & Universal   \\
    \midrule
    \multicolumn{5}{l}{\textit{Image models}}\\
    SD v2.1 & 0.66 & 0.64 & 0.65 & 0.51 & 0.63 & 0.60\\
    SD-XL Turbo & 0.71 & 0.68 & 0.69 & 0.52 & 0.66 & 0.63\\
    SD-XL & 0.72 & 0.70 & 0.69 & 0.50 & 0.67 & 0.63\\
    DeepFloyd-IF & 0.70 & 0.70 & 0.71 & 0.50 & 0.65 & 0.63\\
    Midjourney v6 & 0.77 & 0.77 & 0.76 & 0.50 & 0.73 & 0.68\\
    DALL-E 3 & 0.80 & 0.80 & 0.77 & 0.49 & 0.75 & 0.69\\
    \midrule
    \multicolumn{5}{l}{\textit{Video models}}\\
    ModelScope & 0.58 & 0.61 & 0.57 & 0.52 & 0.52 & 0.55\\
    Floor33 & 0.60 & 0.64 & 0.59 & 0.53 & 0.55 & 0.57\\
    Pika v1 & 0.65 & 0.64 & 0.63 & 0.55 & 0.63 & 0.61\\
    Gen2 & 0.69 & 0.69 & 0.64 & 0.54 & 0.58 & 0.62\\
    \bottomrule[1.1pt]
    \end{NiceTabular} 
    
    \\
  {\bf (c) Human ratings on ``advanced'' prompts}   & {\bf (d) VQAScores on ``advanced'' prompts} 
\end{tabular}
}
\label{tab:performance_breakdown}
\end{table*}

\begin{table}[h!]
\centering
\renewcommand{\arraystretch}{1.3}
\begin{subtable}[t]{\textwidth}
\centering
\scalebox{0.8}{
  \begin{NiceTabular}{lccccc|c}
        \CodeBefore
        \Body
    \toprule[1.2pt]
    \multirow{2}{*}{\textbf{Method}} & \multirow{2}{*}{\bf Attribute} & \multirow{2}{*}{\bf Scene} & \multicolumn{3}{c}{\bf Relation} & \multirow{2}{*}{\bf Overall}   \\
    \cmidrule{4-6}
    &  &  &  Spatial & Action & Part   \\
    \midrule
    \textbf{VQAScore (CLIP-FlanT5)} & \textbf{63.8} & \textbf{62.1} & \textbf{64.5} & \textbf{63.7} & \textbf{65.0} & \textbf{63.7} \\
    \midrule
    BLIP-VQA~\cite{huang2023t2i} & 58.9 & 57.5 & 56.8 & 56.6 & 58.9 & 56.7 \\
    LLMScore~\cite{llmscore} & 54.6 & 56.2 & 55.4 & 57.2 & 54.7 & 55.4 \\
    VQAScore~(InstructBLIP) & 60.9 & 60.3 & 62.2 & 61.9 & 61.9 & 61.1 \\
    VQAScore~(LLaVA-1.5) & 59.4 & 57.5 & 59.2 & 59.8 & 60.3 & 59.0 \\
    Davidsonian~\cite{davidsonian} & 53.5 & 52.9 & 55.1 & 53.7 & 55.3 & 54.2 \\
    VQ2~\cite{vq2} & 51.7 & 51.8 & 52.2 & 52.7 & 53.9 & 52.5 \\
    HPSv2~\cite{hpsv2} & 48.4 & 48.6 & 49.6 & 49.4 & 54.4 & 49.4 \\
    CLIPScore~\cite{clipscore} & 46.8 & 45.1 & 49.3 & 47.9 & 49.4 & 47.4 \\
    BLIPv2Score~\cite{blipv2} & 47.5 & 45.9 & 51.6 & 50.2 & 53.2 & 48.5 \\
    ImageReward~\cite{imagereward} & 55.8 & 55.1 & 58.3 & 57.2 & 59.3 & 56.3 \\
    PickScore~\cite{pickscore} & 57.7 & 57.9 & 57.2 & 58.4 & 59.3 & 58.2 \\
    \textbf{VQAScore~(GPT4-o)} & \textbf{69.3} & \textbf{68.4} & \textbf{69.3} & \textbf{68.1} & \textbf{69.7} & \textbf{69.2} \\
    \bottomrule[1.2pt]
    \end{NiceTabular}
    }
            \caption{Pairwise Accuracy on “basic” skills}
            \label{tab:first_subtable}
        \end{subtable} \\
        
        \vspace{1cm} %

\begin{subtable}[t]{\textwidth}
            \centering   
\scalebox{0.8}{%
    \begin{NiceTabular}{lccccc|c}
        \CodeBefore
        \Body
    \toprule[1.1pt]
    \multirow{2}{*}{\textbf{Method}} &  \multirow{2}{*}{\bf Count} & \multirow{2}{*}{\bf Differ} & \multirow{2}{*}{\bf Compare} & \multicolumn{2}{c}{\bf Logical}  & \multirow{2}{*}{\bf Overall}   \\
    \cmidrule{5-6}
    &  &  & &  Negate & Universal   \\
    \midrule
    \textbf{VQAScore (CLIP-FlanT5)} &\textbf{59.2} & \textbf{58.0} & \textbf{57.3} & \textbf{58.3} & \textbf{59.1} & \textbf{59.6} \\
    \midrule
    BLIP-VQA~\cite{huang2023t2i}               & 55.3 & 49.8 & 51.9 & 44.1 & 54.1 & 51.4 \\
    LLMScore~\cite{llmscore}               & 54.6 & 55.4 & 56.2 & 57.2 & 54.7 & 55.4 \\
    VQAScore~(InstructBLIP)   & 56.8 & 55.1 & 54.8 & 58.6 & 58.2 & 58.5 \\
    VQAScore~(LLaVA-1.5)   & 58.1 & 54.8 & 53.0 & 57.3 & 59.6 & 58.9 \\
    Davidsonian~\cite{davidsonian}            & 53.5 & 55.1 & 52.9 & 53.7 & 55.3 & 54.2 \\
    VQ2~\cite{vq2}                    & 51.7 & 52.2 & 51.8 & 52.7 & 53.9 & 52.5 \\
    HPSv2~\cite{hpsv2}                  & 53.6 & 51.7 & 51.1 & 44.4 & 47.7 & 49.1 \\
    CLIPScore~\cite{clipscore}              & 50.6 & 50.7 & 47.3 & 46.0 & 50.0 & 49.8 \\
    BLIPv2Score~\cite{blipv2}            & 52.3 & 56.5 & 51.5 & 47.1 & 52.3 & 51.6 \\
    ImageReward~\cite{imagereward}            & 56.3 & 55.3 & 56.0 & 46.6 & 52.5 & 53.2 \\
    PickScore~\cite{pickscore}              & 56.4 & 54.0 & 53.9 & 47.9 & 52.4 & 53.7 \\
    \textbf{VQAScore~(GPT4-o) }      &\textbf{ 64.0} &\textbf{ 62.3 }& \textbf{63.0} & \textbf{64.0} & \textbf{68.3} & \textbf{66.2} \\

    \bottomrule[1.1pt]
    \end{NiceTabular} 
}
            \caption{Pairwise Accuracy on “advanced” skills}
            \label{tab:second_subtable}
        \end{subtable}
    \caption{\textbf{Evaluating human correlation across skills on GenAI-Bench (Image).} For a comprehensive comparison, we report the Pairwise Accuracy of different automated metrics across skills. Overall, VQAScore based on our in-house CLIP-FlanT5 or the proprietary GPT4-o surpasses other metrics across all skills by a large margin. Lastly, all automated metrics perform much worse on GenAI-Bench's ``advanced'' prompts that require higher-order reasoning, suggesting room for improvement.}
    \label{tab:comprehenseive_comparison_with_metrics}
\end{table}

\section{Details of Baseline Methods}
\label{sec:method_details}
In this section, we detail the implementation of the baseline methods. \autoref{tab:vqascore_all} reports VQAScore performance on seven more benchmarks that measures correlation with human judgments. 

\begin{table*}[h!]
\centering
\renewcommand{\arraystretch}{1.3}
\caption{\textbf{VQAScore on image-text alignment benchmarks.} We show Group Score for Winoground and EqBen; AUROC for DrawBench, EditBench, and COCO-T2I; pairwise accuracy~\cite{deutsch2023ties} for TIFA160 and GenAI-Bench; and binary accuracy for Pick-a-Pick, with higher scores indicating better performance for all metrics. VQAScore (based on CLIP-FlanT5) outperforms all prior art across all benchmarks. } 
\scalebox{0.6}{
\begin{NiceTabular}{llccccccc|c}
            \CodeBefore
              \rectanglecolor{softgreen}{11-1}{16-10}
            \Body
\toprule[1.5pt]
\multirow{1}{*}{\textbf{Method}}  & \multirow{1}{*}{\textbf{Models}} &  \multirow{1}{*}{{\bf Winoground}} &  \multirow{1}{*}{{\bf EqBen}} &  \multirow{1}{*}{{\bf DrawBench}}  & \multirow{1}{*}{{\bf EditBench}} & \multirow{1}{*}{{\bf COCO-T2I}} &
\multirow{1}{*}{\textbf{TIFA160}} & \multirow{1}{*}{\textbf{Pick-a-Pic}} & \multirow{1}{*}{\textbf{GenAI-Bench}}
\\ 
\midrule
\multicolumn{3}{l}{\textit{{Based on vision-language models}}} \\
\multirow{1}{*}{CLIPScore~\cite{clipscore}}  & CLIP-L-14 & 7.8 & 25.0 & 49.1 & 60.6 & 63.7  & 54.1 & 76.0 & 50.8 \\ \midrule
\multicolumn{3}{l}{\textit{Finetuned on human feedback}} \\
\multirow{1}{*}{PickScore~\cite{pickscore}} & CLIP-H-14 (finetuned) & 6.8 & 23.6 & 72.3 & 64.3 & 61.5 & 59.4  & 70.0 & 56.2 \\
\multirow{1}{*}{ImageReward~\cite{imagereward}} & BLIPv2 (finetuned) & 12.8 & 26.4 & 70.4 & 70.3 & 77.0 & 67.3 & 75.0 & 55.8 \\
\multirow{1}{*}{HPSv2~\cite{hpsv2}} & CLIP-H-14 (finetuned) & 4.0 & 17.0 & 63.1 & 64.1 & 60.3 & 55.2 & 69.0 & 49.6 \\ \midrule
\multicolumn{4}{l}{\textit{QG/A methods}} \\
\multirow{1}{*}{VQ2~\cite{vq2}} & FlanT5, LLaVA-1.5 & 10.0 & 20.0 & 52.8 & 52.8 & 47.7 & 48.7 & 73.0 & 51.9 \\
\multirow{1}{*}{Davidsonian~\cite{davidsonian}} & ChatGPT, LLaVA-1.5 & 15.5 & 20.0 & 78.8 & 69.0 & 76.2 & 54.3 & 70.0 & 54.6 \\ \midrule
\multicolumn{4}{l}{\it VQAScore (ours) using open-source VQA models} \\
\multirow{1}{*}{\bf VQAScore} & InstructBLIP & 28.5 & 38.6 & 82.6 & 75.7 & 83.0 & 70.1 & 83.0 & 62.3 \\ %
\multirow{1}{*}{\bf VQAScore} & LLaVA-1.5 & 29.8 & 35.0 & 82.2 & 70.6 & 79.4 & 66.4 & 76.0 & 61.7\\ %
\midrule
\multicolumn{4}{l}{\textit{VQAScore (ours) using our VQA model}} \\
\multirow{1}{*}{\bf VQAScore} & {\bf CLIP-FlanT5} & {\bf 46.0} & {\bf 47.9} & {\bf 85.3} &	{\bf 77.0} &	{\bf 85.0} &	{\bf 71.2}  &	{\bf 84.0} & {\bf 64.1} \\ %
\bottomrule[1.5pt]
\end{NiceTabular}
}
\label{tab:vqascore_all}
\end{table*}

{\bf CLIPScore and BLIPv2Score.} To calculate CLIPScore, we use the CLIP-L-336 model~\cite{clipscore}. To calculate BLIPv2Score, we use the ITMScore of BLIPv2-vit-G~\cite{blipv2}. 

{\bf Metrics finetuned on human feedback (PickScore/ImageReward/HPSv2).} We use the official code and model checkpoints to calculate these metrics. PickScore~\cite{pickscore} and HPSv2~\cite{hpsv2} finetune the CLIP-H model, and ImageReward~\cite{imagereward} finetunes the BLIPv2~\cite{blipv2}, using costly human feedback from either random web users or expert annotators. These metrics use discriminative pre-trained VLMs, which bottleneck their performance due to bag-of-words encodings. Also, their finetuning datasets may lack compositional texts. Finally, human annotations can be noisy, especially when these annotators are not well trained (e.g., random web users of the Pick-a-pic dataset~\cite{pickscore}).

{\bf QG/A methods (VQ2/Davidsonian).} These divide-and-conquer methods are the most popular in recent text-to-visual evaluation~\cite{dalle3, huang2023t2i, t2vscore, dreamsync}. VQ2~\cite{vq2} uses a finetuned FlanT5 to generate free-form QA pairs and computes the average score of P(answer $|$ image, question). Davidsonian uses a more sophisticated pipeline by prompting ChatGPT to generate yes-or-no QA pairs while avoiding inconsistent questions. For example, given the text ``the moon is over the cow'', if a VQA model already answers ``No'' to ``Is there a cow?'', it then skips the follow-up question ``Is the moon over the cow?''. However, these methods often generate nonsensical QA pairs, as shown in \autoref{tab:divide_errors} on real-world user prompts from GenAI-Bench. 

\begin{table*}[h]
\centering
\caption{\textbf{Failure cases of divide-and-conquer methods (VQ2/Davidsonian).} We show generated question-and-answer pairs of VQ2 and Davidsonian on three GenAI-Bench prompts. These methods often generate irrelevant or erroneous QA pairs (highlighted in \textcolor{red}{red}), especially with more compositional texts.}
\scalebox{0.6}{
\begin{NiceTabular}{M{0.15\linewidth} M{0.6\linewidth} M{0.45\linewidth}}
\CodeBefore
    \Body
\toprule[1.2pt]
 \textbf{Method} & \textbf{Generated questions} & \textbf{Candidate answers (correct answer choice in bold)} \\ \midrule

\multicolumn{3}{c}{Text: ``{\tt a snowy landscape with a cabin, but no smoke from the chimney}''}\\
\multirow{2}{*}{VQ2} & \textcolor{red}{What is the name of the landscape on which it's a cabin?} & {\bf a snowy landscape} \\
  & \textcolor{red}{In this landscape what does the fire not go off?} & {\bf a cabin} \\ \cmidrule{1-3}
\multirow{3}{*}{Davidsonian} & Is there a landscape? & \textbf{yes}, no \\
  & Is there no smoke from the chimney? & \textbf{yes}, no \\ 
  & Is the cabin in the landscape? & \textbf{yes}, no \\ 
  \toprule[1.2pt]

\multicolumn{3}{c}{Text: ``{\tt six people wear white shirts and no people wear red shirts}''}\\
\multirow{2}{*}{VQ2} & \textcolor{red}{What does the average American wear?} & {\bf white shirts} \\
  & \textcolor{red}{What kind of clothes do not all people wear?} & {\bf red shirts} \\ \cmidrule{1-3}
\multirow{3}{*}{Davidsonian} & {Are there people?} & \textbf{yes}, no \\
  & \textcolor{red}{Are the shirts red?} & \textbf{yes}, no \\ 
  & Are the shirts white? & \textbf{yes}, no \\ 
  \toprule[1.2pt]

  \multicolumn{3}{c}{Text: ``{\tt \footnotesize in the classroom there are two boys standing together, the boy in the red jumper is taller than the boy in the white t-shirt}''}\\
\multirow{2}{*}{VQ2} & \textcolor{red}{Where do two tall kids stand?} & {\bf the classroom} \\
  & \textcolor{red}{Which color of jumper is the tallest?} & {\bf the red jumper} \\ \cmidrule{1-3}
\multirow{3}{*}{Davidsonian} & \textcolor{red}{Is the boy in the red jumper wearing a red jumper?} & \textbf{yes}, no \\
  & \textcolor{red}{Is the boy in the white t-shirt wearing a white t-shirt?} & \textbf{yes}, no \\ 
  & Are the boys standing together? & \textbf{yes}, no \\ 

\bottomrule[1.2pt]
\end{NiceTabular}
}
\label{tab:divide_errors}
\end{table*}

\end{document}